\documentclass{article}
\PassOptionsToPackage{numbers}{natbib}

\usepackage[main, final]{neurips_2026}

\usepackage[utf8]{inputenc} 
\usepackage[T1]{fontenc}    
\usepackage{hyperref}       
\usepackage{url}            
\usepackage{booktabs}       
\usepackage{amsfonts}       
\usepackage{nicefrac}       
\usepackage{microtype}      
\usepackage{xcolor}         
\usepackage{amsmath}
\usepackage{graphicx}
\usepackage{multirow}
\usepackage{amssymb}
\usepackage{caption}

\newcommand{\std}[1]{{\scriptsize $\pm$#1}}

\title{When to Trust Memory: \\ Retrieval-Guided Probabilistic Spatiotemporal Forecasting under Distribution Shift}

\author{
  Haochen Lv\textsuperscript{1} \\
  \texttt{lvhaochen@bjtu.edu.cn}
  \And
  Jianhao Zhang\textsuperscript{1} \\
  \texttt{jianhaozhang@bjtu.edu.cn}
  \And
  Zhichen Lei\textsuperscript{1} \\
  \texttt{leizhichen@bjtu.edu.cn}
  \AND
  Yang Jiang\textsuperscript{1} \\
  \texttt{jiangyang@bjtu.edu.cn}
  \And
  Xiaowei Mao\textsuperscript{1} \\
  \texttt{maoxiaowei@bjtu.edu.cn}
  \And
  Shengnan Guo\textsuperscript{1,2,*} \\
  \texttt{guoshn@bjtu.edu.cn}
  \AND
  Youfang Lin\textsuperscript{1,3} \\
  \texttt{yflin@bjtu.edu.cn}
  \And
  Huaiyu Wan\textsuperscript{1,3} \\
  \texttt{hywan@bjtu.edu.cn}
}
\begin{document}

\maketitle

\begingroup
\makeatletter
\renewcommand{\thefootnote}{}
\renewcommand{\@makefntext}[1]{\noindent #1}
\makeatother

\begin{NoHyper}
\footnotetext{
\textsuperscript{*} Corresponding author \\
\textsuperscript{1} School of Computer Science and Technology,
Beijing Jiaotong University, Beijing, China. \\
\textsuperscript{2} Key Laboratory of Big Data and Artificial Intelligence
in Transportation, Ministry of Education, Beijing, China. \\
\textsuperscript{3} Beijing Key Laboratory of Traffic Data Mining
and Embodied Intelligence, Beijing, China. 
}
\end{NoHyper}
\endgroup

\begin{abstract}
  Spatiotemporal forecasting plays an important role in real-world applications, where probabilistic approaches provide uncertainty estimates for decision-making. Real-world systems are inherently non-stationary, leading to distribution shifts between training and test data. However, existing probabilistic methods mainly focus on modeling data distributions under standard settings, while distribution shift methods often handle such shifts in a coarse manner, failing to characterize shift degree and adapt prediction strategy accordingly. As a result, they often produce miscalibrated and unreliable forecasts. To address this issue, we propose REMIND, a retrieval-guided probabilistic framework for spatiotemporal forecasting under distribution shift. REMIND leverages retrieval to estimate how well current observation are supported by historical patterns, enabling adaptive prediction under different shift regimes. It employs spatiotemporal retrieval to construct a memory bank and introduces a dual branch architecture, combining a memory-calibrated probabilistic branch for mild shifts with a deviation-aware probabilistic branch for severe shifts. A shift-adaptive gate, guided by retrieval similarity and memory dispersion, balances the two branches according to shift degree. We further design a hierarchical Gaussian mixture to capture heterogeneous uncertainty over the dual branch predictions, and introduce a memory perturbation training scheme to improve generalization. Experiments on real-world datasets with distribution shifts demonstrate that REMIND consistently outperforms state-of-the-art methods in both forecasting accuracy and probabilistic prediction quality. Code is available at \url{https://github.com/LvHaochenBANG/REMIND.git}.

\end{abstract}

\vspace{-2mm}
\section{Introduction}
\vspace{-2mm}
  Spatiotemporal forecasting plays a crucial role in real-world applications, such as transportation~\cite{guo2023self,guo2021learning,shen2025towards}, hydrology~\cite{rentschler2022flood}, and energy systems~\cite{zhang2025strap,11002729, cheng2026metagnsdformer}. In recent years, probabilistic approaches have gained increasing attention, as they aim to model the underlying distribution of system dynamics~\cite{salinas2020deepar,an2024spatio,wen2023diffstg}. Beyond point predictions, these methods provide uncertainty quantification, which is essential for risk assessment, decision-making, and early warning~\cite{wu2021quantifying, jin2024spatial, lv2025ripcn, yu2025merlin}.
  
  Most existing probabilistic spatiotemporal forecasting methods focus on modeling data distributions under standard settings. However, real-world spatiotemporal systems are inherently non-stationary, leading to distribution shifts, reflected as a mismatch between training and test data distributions~\cite{sun2024learning}. Such shifts commonly arise from evolving system dynamics, including changing patterns, sensor noise, and unexpected events such as accidents or extreme weather~\cite{liu2022non, lim2021temporal}. Importantly, distribution shift is inherently continuous rather than binary~\cite{zhou2022domain}. At one end, the absence of shift corresponds to in-distribution (ID) scenarios, while at the other end, extreme shifts correspond to out-of-distribution (OOD) scenarios~\cite{li2025out}. In practice, many real-world cases lie between these two extremes.

  Recently, several methods have been proposed to handle distribution shift. ShifTS~\cite{zhao2025tackling} improves generalization by learning invariant patterns of input features under different external conditions. TFPS~\cite{sun2024learning} performs adaptive forecasting by dynamically selecting experts specialized in specific patterns. In particular, retrieval-based approaches such as RAFT~\cite{han2025retrieval} and STRAP~\cite{zhang2025strap} have shown strong potential by leveraging an external memory bank of historical data to retrieve and preserve diverse patterns, especially rare ones, thereby improving predictive performance across diverse distributional conditions. 
  
  Despite these advances, handling distribution shift remains challenging in practice. Existing methods often handle distribution shift in a coarse manner, lacking the ability to explicitly characterize shift degree and adapt prediction strategies accordingly, resulting in miscalibrated predictive distributions and unreliable forecasts. Retrieval-based methods, for example, typically retrieve historical samples based on historical observations, rank them by similarity, and reuse their associated future outcomes, implicitly assuming that the retrieved samples can be trusted without assessing their reliability. However, this assumption may fail in two ways. First, under severe distribution shifts, even the most similar historical samples may differ substantially from the current observation, providing limited support for prediction. Second, similar historical observations may correspond to markedly different future outcomes, making input similarity alone insufficient to determine the reliability of retrieved information. Consequently, reusing retrieved samples without assessing their relevance and consistency may bias predictions and undermine uncertainty estimation.
  
  To address these challenges, we propose REMIND, a retrieval-guided probabilistic framework for spatiotemporal forecasting under distribution shift. Unlike conventional approaches that treat retrieval as passive memory lookup, REMIND leverages retrieval as an active signal of how well the current observation is supported by historical patterns. Specifically, it assesses both the relevance of retrieved observations and the consistency of their associated future outcomes to guide adaptive probabilistic forecasting.
  
  Concretely, we design a coarse-to-fine spatiotemporal retrieval mechanism that equips REMIND with a memory bank and achieves both high retrieval quality and efficiency. Based on the retrieval, we construct two complementary branches. A memory-calibrated probabilistic branch predicts from retrieved patterns through lightweight calibration and is effective under mild shifts. A deviation-aware probabilistic branch predicts from the current observation and uses deviations from retrieved patterns as residual corrections, making it more suitable for severe shifts. REMIND then fuses the two branches via a shift-adaptive gate, which leverages retrieval similarity and memory dispersion to estimate shift degree and balance them accordingly.
  In addition, we design a hierarchical Gaussian mixture to capture heterogeneous uncertainty through inner-branch uncertainty and cross-branch disagreement, and introduce a memory perturbation training scheme to improve generalization under distribution shift.

  Our contributions are summarized as follows:
  \begin{itemize}
    \item We propose REMIND, a retrieval-guided probabilistic framework for spatiotemporal forecasting under distribution shift. By treating retrieval as an active signal of support from historical patterns, it adaptively balances a memory-calibrated probabilistic branch and a deviation-aware probabilistic branch according to shift degree, enabling adaptive prediction across different shift regimes.
    \item We develop a hierarchical Gaussian mixture to model heterogeneous predictive uncertainty over the dual branch predictions, and introduce a memory perturbation training scheme to improve generalization under distribution shift.
    \item Experiments on real-world datasets from hydrology, traffic and energy domains with spatiotemporal distribution shifts demonstrate that our method consistently outperforms state-of-the-art approaches in both forecasting accuracy and probabilistic prediction quality.
  \end{itemize}

\vspace{-2mm}
\section{Related Work}
\vspace{-2mm}
Probabilistic spatiotemporal forecasting models the predictive distribution of future system states. Existing methods can be broadly divided into parametric distribution models and generative distribution models. Parametric distribution methods assume a predefined distribution family, such as Gaussian distributions, and learn its parameters, as in DeepAR~\cite{salinas2020deepar}, DER~\cite{amini2020deep}, SDER~\cite{wu2024evidence}, and LODE~\cite{rubanova2019latent}. They are computationally efficient but limited by their distributional assumptions. In contrast, generative distribution models, including diffusion-based methods such as PriSTI~\cite{liu2023pristi} and DiffSTG~\cite{wen2023diffstg} and flow-based methods such as STGNF~\cite{an2024spatio} and STMPN~\cite{an2025spatio}, learn complex distributions without fixed forms. While more expressive, they typically rely on repeated sampling, resulting in higher computational cost. These methods mainly focus on modeling data distributions under standard settings, where distribution shift is not explicitly considered. In practice, spatiotemporal systems are often dynamic and non-stationary, and distribution changes may occur over time, which can affect predictive calibration.

To improve generalization under such conditions, several methods have been proposed to mitigate the effects of distribution shift. ShifTS~\cite{zhao2025tackling} learns invariant representations under varying conditions to enhance generalization across shifting environments, while TFPS~\cite{sun2024learning} dynamically selects specialized experts to adapt to diverse temporal patterns. STONE~\cite{wang2024stone} captures invariant node dependencies through gated spatiotemporal semantic graphs and graph intervention, thereby improving generalization under spatiotemporal distribution shifts. Retrieval-based methods, such as RAFT~\cite{han2025retrieval}, RATD~\cite{liu2024retrieval}, and STRAP~\cite{zhang2025strap}, further enhance generalization by leveraging historical memory, allowing models to capture diverse patterns and recall rare or previously observed events. While these approaches have shown promising results, several challenges remain. Specifically, existing methods often do not explicitly characterize the degree of distribution shift and adapt prediction strategy accordingly, resulting in miscalibrated predictive distributions. In addition, these approaches are primarily deterministic in nature and focus on adjusting point estimates rather than modeling the full predictive distribution. 

\vspace{-1mm}
\section{Preliminaries}
\vspace{-2mm}
\textbf{Spatial Graph.}
We model the spatial structure using a graph $\mathcal{G} = (\mathcal{V}, \mathcal{E}, \boldsymbol{A})$, where $\mathcal{V}$ is the set of $N$ nodes representing spatial locations, $\mathcal{E}$ is the set of edges capturing the relationships between nodes, and $\boldsymbol{A} \in \mathbb{R}^{N \times N}$ denotes the adjacency matrix.

\textbf{Probabilistic Spatiotemporal Forecasting.}
At each time step $t$, let $x_{t,n} \in \mathbb{R}^{F}$ denote the observation at location $n$ with $F$ variables. Aggregating observations from all locations gives $\boldsymbol{X}_t = [x_{t,1}, x_{t,2}, \dots, x_{t,N}]^\top \in \mathbb{R}^{N \times F}$, representing multivariate observations over $N$ locations. We use $\boldsymbol{X}_H$ to denote the historical sequence over the past $T$ time steps, corresponding to $\boldsymbol{X}_{t-T+1:t} \in \mathbb{R}^{T \times N \times F}$, and use $\boldsymbol{X}_P$ to denote the future sequence over the next $T$ time steps, corresponding to $\boldsymbol{X}_{t+1:t+T} \in \mathbb{R}^{T \times N \times F}$.
Given $\boldsymbol{X}_H$, the goal is to learn a probabilistic mapping $\mathcal{F}$ that characterizes the distribution of $\boldsymbol{X}_P$ and produces its predictive statistics, i.e., $\left( \boldsymbol{\mu}_{P}, \boldsymbol{\sigma}_{P} \right) = \mathcal{F}\left( \boldsymbol{X}_{H}, \mathcal{G} \right)$, where $\boldsymbol{\mu}_{P}, \boldsymbol{\sigma}_{P} \in \mathbb{R}^{T \times N \times F}$.

\vspace{-1mm}
\section{REMIND}
\vspace{-2mm}
\begin{figure}[t]
  \centering
  \makebox[\linewidth][c]{%
  \includegraphics[width=1.05\linewidth]{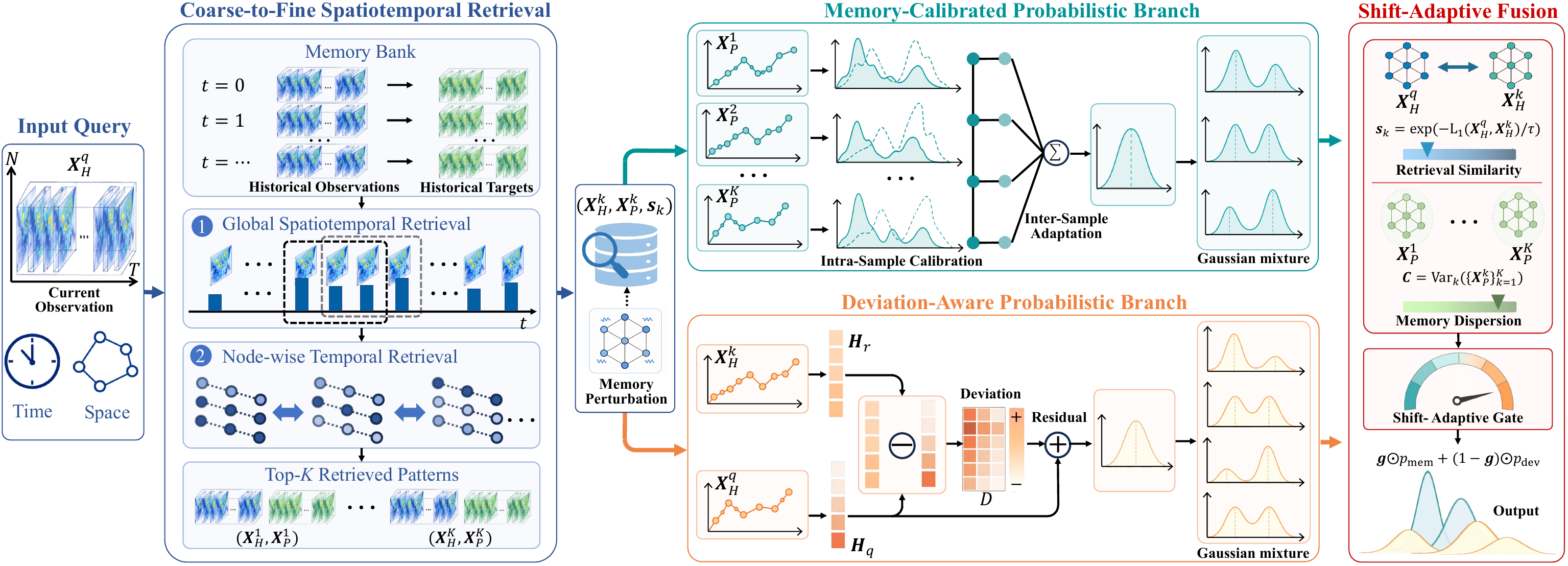}
  }
  \caption{The overall architecture of REMIND. REMIND performs coarse-to-fine spatiotemporal retrieval to construct a memory bank and employs a dual branch architecture for probabilistic forecasting under distribution shift. A shift-adaptive gate dynamically balances the branches based on retrieval similarity and memory dispersion, and a hierarchical Gaussian mixture models heterogeneous predictive distribution.}
  \label{overview}
  \vspace{-5mm}
\end{figure}

\subsection{Overview}
The key idea of REMIND is to treat retrieval similarity not only as auxiliary evidence for prediction, but also as a signal indicating how well the current observation is supported by historical patterns. REMIND assesses both the relevance of retrieved observations and the consistency of their associated future outcomes, enabling adaptive memory utilization under distribution shift. Intuitively, strong historical support suggests that retrieved patterns are both relevant and informative, allowing their associated targets to be reused. In contrast, weak historical support may arise from substantial deviations from historical observations or inconsistent future outcomes among similar patterns. In such cases, the model reduces its reliance on memory, predicts primarily from the current observation, and uses deviations from retrieved patterns as residual corrections.

As shown in~Fig.\ref{overview}, REMIND constructs a memory bank through coarse-to-fine spatiotemporal retrieval and employs a dual branch architecture consisting of a memory-calibrated probabilistic branch and a deviation-aware probabilistic branch. The memory-calibrated probabilistic branch performs lightweight calibration on retrieved targets, while the deviation-aware probabilistic branch models residual deviations from retrieved patterns. A shift-adaptive gate balances the two branches according to retrieval similarity and memory dispersion. Finally, REMIND outputs a hierarchical Gaussian mixture distribution to capture heterogeneous uncertainty under different shift regimes.

\vspace{-2mm}
\subsection{Coarse-to-Fine Spatiotemporal Retrieval}
\vspace{-2mm}
Let $\mathcal{D} = \{(\boldsymbol{X}_H^{i}, \boldsymbol{X}_P^{i})\}_{i=1}^R$ denote a memory bank of $R$ historical observation-target pairs. In an ideal spatiotemporal retrieval mechanism, given a query observation $\boldsymbol{X}_H^{q}$, the system aims to rapidly and efficiently retrieve the top $K$ most similar samples $\mathcal{D}_{\mathrm{ret}} = \{\boldsymbol{X}_H^{k}, \boldsymbol{X}_P^{k}, \boldsymbol{s}_k\}_{k=1}^K$, where $\boldsymbol{s}_k$ denotes the similarity score between $\boldsymbol{X}_H^{q}$ and $\boldsymbol{X}_H^{k}$.

Retrieving spatiotemporal data is non-trivial due to the entanglement of spatial and temporal dimensions. Accordingly, spatiotemporal retrieval can be formulated at four levels: global spatiotemporal matching, node-wise temporal matching, time-wise spatial matching, and point-wise matching. Global spatiotemporal matching preserves overall structure with high efficiency but enforces strict temporal alignment across nodes. In contrast, node-wise temporal matching retrieves patterns independently for each node based on its temporal dynamics, enabling more flexible and fine-grained retrieval, but requiring independent matching for each node and thus increasing computational cost. Meanwhile, time-wise spatial matching disrupts temporal continuity, and point-wise matching ignores global spatiotemporal structure, limiting their ability to capture coherent spatiotemporal patterns (see App.~\ref{app:retrieval} for further discussion).

To balance retrieval quality and efficiency, we propose a coarse-to-fine spatiotemporal retrieval framework consisting of global spatiotemporal retrieval followed by node-wise temporal retrieval.

In the first stage, we perform global spatiotemporal retrieval over $\mathcal{D}$, where samples are constructed via sliding windows with a stride of 1 along the temporal dimension. Treating the current observation $\boldsymbol{X}_H^{q} \in \mathbb{R}^{T \times N \times F}$ as a whole, we select a candidate set $\mathcal{C} = \{(\boldsymbol{X}_H^{k}, \boldsymbol{X}_P^{k}, \boldsymbol{s}_k)\}_{k=1}^{\tilde{K}}$ based on global spatiotemporal similarity. This stage narrows the search space to historically relevant samples with similar overall spatiotemporal patterns.

In the second stage, we perform node-wise temporal retrieval within the candidate set $\mathcal{C}$. Specifically, for each query node $n$, we compare its temporal pattern $\boldsymbol{X}_{H,n}^{q}$ with the candidate nodes drawn from a node set $\mathcal{N}(n)$, which includes both physically connected neighbors and functionally similar nodes defined by Pearson correlation. The cross-spatial similarity can be written as
\begin{equation}
  s_{k,n,m} = \mathrm{Sim}\!\left(\boldsymbol{X}_{H,n}^{q}, \boldsymbol{X}_{H,m}^{k}\right), \quad m \in \mathcal{N}(n).
\end{equation}
Based on these similarities, the refined retrieval results can be expressed as $\mathcal{D}_{\mathrm{ret}}^{(n)} \subseteq \{\boldsymbol{X}_{H,m}^{k}, \boldsymbol{X}_{P,m}^{k}, s_{k,n,m}\}_{k=1}^{K}$. After processing all nodes, we obtain the final retrieval set $\mathcal{D}_{\mathrm{ret}} = \{\boldsymbol{X}_H^{k}, \boldsymbol{X}_P^{k}, \boldsymbol{s}_k\}_{k=1}^K$. Compared with global retrieval, this stage allows the model to capture local temporal misalignment commonly observed under distribution shift. It also leverages similar dynamics from correlated neighboring nodes, thereby enriching retrieval diversity. 

The similarity score for the $k$-th retrieved sample is defined as
\begin{equation}
\boldsymbol{s}_k = \exp\left(-\mathrm{L}_1(\boldsymbol{X}_H^q, \boldsymbol{X}_H^{k}) / \tau \right),
\end{equation}
where $\mathrm{L}_1(\cdot,\cdot)$ denotes the $\ell_1$ distance and $\tau$ is a temperature parameter that controls the concentration of the similarity scores~\cite{zhang2025strap}. This formulation captures overall similarity and supports fully parallel computation of similarities between $\boldsymbol{X}_H^q$ and the entire memory bank.

\vspace{-1mm}
\subsection{Memory-Calibrated Probabilistic Branch}
\vspace{-1mm}
\label{Memory-based Branch}
The memory-calibrated probabilistic branch is designed for scenarios where retrieved patterns remain reliable under mild distribution shifts. In such cases, the current observation is still well supported by historical patterns, allowing retrieved targets to provide informative future references. Therefore, instead of predicting future dynamics from scratch, this branch directly leverages retrieved targets and performs lightweight query-conditioned calibration to adapt them to current observations.

We first encode the current observation and the $k$-th retrieved observation:
\begin{equation}
\boldsymbol{H}_q=E_q(\boldsymbol{X}_H^q), \qquad \boldsymbol{H}_{r}^k=E_r(\boldsymbol{X}^{k}_H),
\end{equation}
where $\boldsymbol{H}_q, \boldsymbol{H}_{r}^k \in \mathbb{R}^{N \times T \times d}$. $E_q$ and $E_r$ are lightweight spatiotemporal encoders composed of GCNs~\cite{zhou2020graph} and temporal attention~\cite{vaswani2017attention}. 

Next, for each retrieved target $\boldsymbol{X}_P^{k}$, we apply a lightweight calibration:
\begin{equation}
\tilde{\boldsymbol{X}}^{k}_P = \boldsymbol{a}_k \odot \boldsymbol{X}^{k}_P + \boldsymbol{b}_k,
\end{equation}
where $\boldsymbol{a}_k$ and $\boldsymbol{b}_k$ are learnable parameters, and $\odot$ denotes element-wise multiplication. To improve stability and prevent overfitting, the parameters are modeled in a low-rank separable form~\cite{kolda2009tensor}. Specifically, $\boldsymbol{a}_k$ is constructed as the outer product of a spatial factor $\boldsymbol{a}_k^{\text{node}} \in \mathbb{R}^{N}$ and a temporal factor $\boldsymbol{a}_k^{\text{time}} \in \mathbb{R}^{T}$, i.e., $\boldsymbol{a}_k = \boldsymbol{a}_k^{\text{node}} (\boldsymbol{a}_k^{\text{time}})^\top$, with $\boldsymbol{b}_k$ defined analogously.

In addition to the above intra-sample calibration, we further perform inter-sample adaptation. We compute a relevance weight for each retrieved sample:
\begin{equation}
  \boldsymbol{\alpha}_k
  = \mathrm{softmax}\!\left(
  \boldsymbol{s}_k + d^{-1/2}\,
  \langle \bar{\boldsymbol{H}}_q,\bar{\boldsymbol{H}}_r^k\rangle
  \right),
\end{equation}
where $\bar{\boldsymbol{H}}_q, \bar{\boldsymbol{H}}_r^k \in \mathbb{R}^{N \times d}$ are temporally aggregated representations obtained via temporal convolution~\cite{bai2018empirical} followed by pooling. The Frobenius inner product $\langle \cdot, \cdot \rangle$ quantifies the similarity between the query and retrieved representations, while $\boldsymbol{s}_k$ provides a retrieval-based prior. Unlike directly normalizing retrieval similarity, the query-conditioned weighting is not solely determined by historical similarity, allowing the model to adaptively adjust the contributions of retrieved samples. Using these relevance weights, the base predictive mean of memory-calibrated probabilistic branch is computed as a weighted aggregation of calibrated targets $\boldsymbol{\mu}_{\mathrm{mem}} = \sum_{k=1}^{K}\boldsymbol{\alpha}_k \tilde{\boldsymbol{X}}_{P}^{k}$.

To capture uncertainty, we relate the predictive variance to both representation features and memory dispersion $\boldsymbol{C} = \mathrm{Var}_{k}(\{{\boldsymbol{X}}_P^{k}\}_{k=1}^{K}) \in \mathbb{R}^{T \times N \times F}$. Intuitively, larger dispersion indicates greater variability among retrieved targets, implying higher uncertainty over possible future outcomes. Based on this, the base variance is estimated as:
\begin{equation}
\boldsymbol{\sigma}_{\text{mem}}=\mathrm{Softplus}\big(f_{\sigma}(\boldsymbol{H}_q,{\boldsymbol{H}}_r)+f_c(\boldsymbol{C})\big),
\end{equation}
where ${\boldsymbol{H}}_r = \sum_{k=1}^{K} \boldsymbol{\alpha}_k \boldsymbol{H}_r^{k}$ denotes the aggregated retrieved representation, $f_{\sigma}$  and $f_c$ are implemented as lightweight MLPs.

To better capture the inherent heterogeneity of spatiotemporal forecasting, we parameterize the predictive distribution of the memory-calibrated probabilistic branch as a Gaussian mixture~\cite{bishop2006pattern}:
\begin{equation}
p_{\mathrm{mem}}(\boldsymbol{X}_P\mid \boldsymbol{X}_H)
=
\sum_{i=1}^{I_{\mathrm{mem}}}
\boldsymbol{\pi}_i\;
\mathcal{N}\!\left(
\boldsymbol{X}_P;
\boldsymbol{\mu}_{\mathrm{mem}} + \Delta \boldsymbol{\mu}_{i},
\left(\boldsymbol{\sigma}_{\mathrm{mem}} + \Delta \boldsymbol{\sigma}_{i}\right)^2
\right),
\label{eq:mem_mg}
\end{equation}
where the base terms $\boldsymbol{\mu}_{\mathrm{mem}}$ and $\boldsymbol{\sigma}_{\mathrm{mem}}$ provide a retrieval-based prior, while the residual terms $\Delta \boldsymbol{\mu}_{i}$ and $\Delta \boldsymbol{\sigma}_{i}$ enable flexible refinement for each component. The mixture weights $\boldsymbol{\pi}_i$ and the residual parameters ${\Delta \boldsymbol{\mu}_{i}, \Delta \boldsymbol{\sigma}_{i}}$ are predicted from the memory context $(\boldsymbol{H}_q, {\boldsymbol{H}}_r)$. In addition to providing expressive distribution modeling, the mixture formulation also facilitates the subsequent probabilistic fusion between branches, since the gate-weighted combination of branch-wise mixtures naturally forms a unified hierarchical Gaussian mixture distribution.

\vspace{-2mm}
\subsection{Deviation-Aware Probabilistic Branch}
\vspace{-1mm}
\label{Deviation-aware Branch}
The deviation-aware probabilistic branch is designed for scenarios where retrieved patterns become less reliable under severe distribution shifts. In such cases, retrieved targets may no longer accurately represent future evolution, while deviations between current observations and historical patterns become informative for characterizing shift dynamics. Therefore, instead of directly reusing retrieved targets, this branch predicts from current observations and leverages deviations from retrieved patterns as residual corrections.

Similar to the memory-calibrated probabilistic branch, we encode the current observation and retrieved observations into spatiotemporal representations $\boldsymbol{H}_q$ and $\{\boldsymbol{H}_{r}^{k}\}_{k=1}^{K}$, via an encoder which is independent of that in the memory-calibrated probabilistic branch. The retrieved representations are then aggregated by similarity weights as $\boldsymbol{H}_r = \sum_{k=1}^{K} \boldsymbol{s}_k \boldsymbol{H}_r^{k}$, yielding a reference representation that summarizes historical patterns relevant to the current observation. 

To quantify the deviation between the current observation and the retrieved reference representation, we construct a deviation representation based on their feature differences and interactions. Specifically, we define
\begin{equation}
\boldsymbol{D}=\phi([\boldsymbol{H}_q-\boldsymbol{H}_r,\; \boldsymbol{H}_q \odot \boldsymbol{H}_r]),
\label{eq:deviation}
\end{equation}
where $[\cdot,\cdot]$ denotes concatenation, and $\phi(\cdot)$ is implemented as MLPs that maps the combined features to a compact deviation representation. Based on $\boldsymbol{D}$, the base predictive mean and variance of the deviation-aware probabilistic branch are defined as
\begin{equation}
  \boldsymbol{\mu}_{\mathrm{dev}} = f_{\mu}(\boldsymbol{H}_q) + f_{\mu}^d(\boldsymbol{D}), \quad
  \boldsymbol{\sigma}_{\mathrm{dev}} = \mathrm{Softplus}\!\left(f_{\sigma}(\boldsymbol{H}_q)\right) + f_{\sigma}^d(\boldsymbol{D}),
\end{equation}
where $f_{\mu}(\boldsymbol{H}_q)$ and $f_{\sigma}(\boldsymbol{H}_q)$ model the primary prediction from the current observation, while $f_{\mu}^d(\boldsymbol{D})$ and $f_{\sigma}^d(\boldsymbol{D})$ provides deviation-aware adjustments. This residual design enables the model to adapt its prediction and uncertainty according to deviations from historical patterns.

Finally, similar to the memory-calibrated probabilistic branch, the deviation-aware probabilistic branch also outputs a Gaussian mixture distribution:
\begin{equation}
  p_{\mathrm{dev}}(\boldsymbol{X}_P\mid \boldsymbol{X}_H)
  =
  \sum_{i=1}^{I_{\mathrm{dev}}}
  \boldsymbol{\pi}_i\;
  \mathcal{N}\!\left(
  \boldsymbol{X}_P;
  \boldsymbol{\mu}_{\mathrm{dev}} + \Delta \boldsymbol{\mu}_{i},
  \left(\boldsymbol{\sigma}_{\mathrm{dev}} + \Delta \boldsymbol{\sigma}_{i}\right)^2
  \right).
\label{eq:dev_mg}
\end{equation}
Compared with the memory-calibrated probabilistic branch, the deviation-aware probabilistic branch requires more mixture components because it cannot directly rely on retrieved targets to provide structured future patterns. Under severe distribution shift, the model must infer possible future evolutions from deviations between current observations and historical patterns, which generally introduces greater ambiguity in future evolution. Additional mixture components therefore improve the capacity of the branch to model heterogeneous uncertainties and diverse future outcomes.

\subsection{Shift-Adaptive Probabilistic Fusion}
\label{Trust Gate}
\begin{figure}[t]
  \centering
  \makebox[\linewidth][c]{%
  \includegraphics[width=1.1\linewidth]{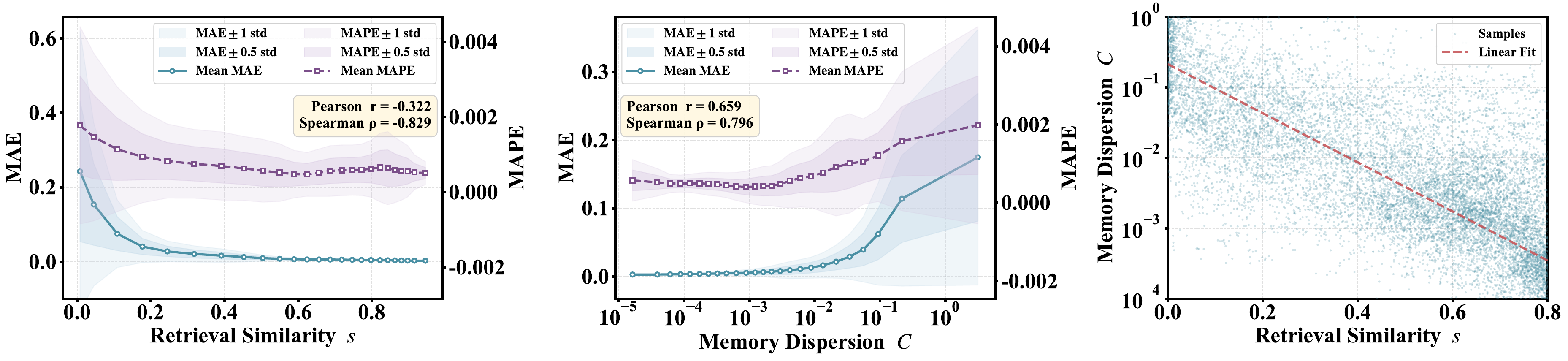}
  }
  \caption{Empirical analysis of retrieval similarity and memory dispersion under distribution shift on the CAMELS dataset. We analyze their relationships with error (MAE/MAPE) between retrieved targets and ground-truth future targets, error variance, and their pairwise correlation.}
  \label{data_stat2}
  \vspace{-5mm}
\end{figure}

Since the two probabilistic branches of REMIND are designed for different shift regimes, their contributions should adapt according to the degree of distribution shift. To achieve this, we introduce a shift-adaptive probabilistic fusion module that dynamically balances the two branches based on retrieval similarity $\boldsymbol{s}$ and memory dispersion $\boldsymbol{C}$. These signals provide complementary indicators of distribution shift and guide adaptive probabilistic fusion under varying shift conditions.

Intuitively, retrieval similarity $\boldsymbol{s}$ reflects how well the current observation is supported by historical patterns. Higher similarity generally indicates that the current observation remains closer to the training distribution, making retrieved targets more reliable for prediction. In contrast, memory dispersion $\boldsymbol{C}$ characterizes the variability among retrieved targets. Even when retrieved observations are historically similar, their corresponding targets may still exhibit diverse future evolutions, resulting in larger dispersion. This suggests weaker consensus in historical support and potentially greater uncertainty under distribution shift. Therefore, $\boldsymbol{s}$ and $\boldsymbol{C}$ provide complementary signals for characterizing distribution shift and guiding adaptive probabilistic fusion.

From a statistical perspective, the historical observation-target pairs in the memory bank can be viewed as samples from a joint distribution, which can be factorized as
$p(\boldsymbol{X}_H,\boldsymbol{X}_P)=p(\boldsymbol{X}_H)p(\boldsymbol{X}_P\mid \boldsymbol{X}_H)$.
Retrieval similarity \(\boldsymbol{s}\) can indicate how well the current observation aligns with historical samples from the marginal distribution \(p(\boldsymbol{X}_H)\), while memory dispersion \(\boldsymbol{C}\) can reflect the variability of the conditional distribution \(p(\boldsymbol{X}_P\mid\boldsymbol{X}_H)\) in the neighborhood of the current observation. These two statistics thus provide complementary evidence for guiding adaptive probabilistic fusion.

Empirical analysis further confirms the effectiveness of retrieval similarity $\boldsymbol{s}$ and memory dispersion $\boldsymbol{C}$ as indicators of distribution shift, as shown in Fig.~\ref{data_stat2}. Overall, lower similarity and larger dispersion are associated with larger forecasting errors and larger error variance, indicating weaker historical support under distribution shift. Moreover, the relationship between the two signals remains relatively dispersed, suggesting that they capture different aspects of distribution shift. These observations support the use of both $\boldsymbol{s}$ and $\boldsymbol{C}$ for adaptive probabilistic fusion.

Motivated by the above observations, we compute the shift-adaptive gate:
\begin{equation}
\boldsymbol{g}=\mathrm{Sigmoid}\left(f_g([\bar{\boldsymbol{s}}, \boldsymbol{s}_1, \boldsymbol{C}, \boldsymbol{D}])\right),
\end{equation}
where $\bar{\boldsymbol{s}}$ and $\boldsymbol{s}_1$ denote the average and top-$1$ retrieval similarities, respectively. $\boldsymbol{C}$ is defined in Sec.\ref{Memory-based Branch}, while deviation $\boldsymbol{D}$ is defined by Eq.\ref{eq:deviation}. Here, $\boldsymbol{D}$ complements $\boldsymbol{s}$ by modeling deviation patterns between current observations and historical patterns.

To guide the shift-adaptive gate toward shift-consistent behavior, we impose soft regularization using $\boldsymbol{s}$ and $\boldsymbol{C}$. First, we encourage the gate $\boldsymbol{g}$ to increase with higher retrieval similarity $\boldsymbol{s}$:
\begin{equation}
  \mathcal{L}_{\mathrm{sim}} = - \mathbb{E} \left[ \boldsymbol{s} \log \boldsymbol{g} + (1 - \boldsymbol{s}) \log (1 - \boldsymbol{g}) \right].
\label{eq:regular}
\end{equation}
For memory dispersion $\boldsymbol{C}$, we introduce a complementary regularization term $\mathcal{L}_{\mathrm{dis}}$, which shares the same form as $\mathcal{L}_{\mathrm{sim}}$ and encourages lower gate values under larger dispersion. These regularizations provide soft directional guidance for gate optimization rather than enforcing hard constraints.

Based on the learned shift-adaptive gate $\boldsymbol{g}$, the final predictive distribution is obtained via a weighted combination of the branch-wise Gaussian mixtures defined in Eqs.\ref{eq:mem_mg} and\ref{eq:dev_mg}:
\begin{equation}
  p(\boldsymbol{X}_P|\boldsymbol{X}_H)= \boldsymbol{g} \odot p_{\text{mem}}(\boldsymbol{X}_P|\boldsymbol{X}_H) + (1-\boldsymbol{g}) \odot p_{\text{dev}}(\boldsymbol{X}_P|\boldsymbol{X}_H).
\end{equation}
This formulation establishes a hierarchical Gaussian mixture, enabling REMIND to jointly model intra-branch multimodality and cross-branch adaptive fusion. From a probabilistic perspective, the resulting predictive variance can be decomposed as
\begin{equation}
  \mathrm{Var}[\boldsymbol{X}_P]
  = \boldsymbol{g} \, \bar{\boldsymbol{\sigma}}_{\mathrm{mem}}^2
  + (1-\boldsymbol{g}) \, \bar{\boldsymbol{\sigma}}_{\mathrm{dev}}^2
  + \boldsymbol{g}(1-\boldsymbol{g}) \, \big(\bar{\boldsymbol{\mu}}_{\mathrm{mem}} - \bar{\boldsymbol{\mu}}_{\mathrm{dev}}\big)^{2},
  \label{eq:vardecomp}
\end{equation}
where $\bar{(\cdot)}$ denotes the branch-wise aggregated means and variances of the corresponding Gaussian mixtures. This formulation shows that the final predictive uncertainty depends on both inner-branch uncertainty, characterized by the first two terms, and cross-branch disagreement, captured by the last term. Detailed derivation is provided in App.~\ref{app:hgmm}.

\subsection{Training Objective and Memory Perturbation}
The model is trained using the negative log-likelihood (NLL) of the predictive distribution:
\begin{equation}
  \mathcal{L}_{\mathrm{NLL}}
  =
  - \mathbb{E}
  \left[
  \log p(\boldsymbol{X}_P \mid \boldsymbol{X}_H)
  \right].
\end{equation}
We further derive the explicit form of the negative log-likelihood under the hierarchical Gaussian mixture by expanding the two-level mixture into a unified mixture (see App.~\ref{app:hgmm}).

Following the gate regularization terms introduced in Eq.~\ref{eq:regular}, the overall objective is defined as:
\begin{equation}
  \mathcal{L}=\mathcal{L}_{\text{NLL}}+\lambda(\mathcal{L}_{\text{sim}}+\mathcal{L}_{\text{dis}}).
\end{equation}

To further improve generalization under distribution shift, we adopt a memory perturbation training scheme to prevent the model from over-relying on highly confident retrievals. Specifically, we randomly attenuate the memory gate as $\tilde{\boldsymbol{g}} = \rho \cdot \boldsymbol{g}$ with $\rho \sim \mathrm{Uniform}(1-p, 1)$, which serves as a soft memory dropout. Here, $p \in [0,1]$ controls the degree of attenuation. In addition, we perturb the retrieved observations and targets as
\begin{equation}
  \tilde{\boldsymbol{X}}_H^{k} = \boldsymbol{X}_H^{k} + (1-\rho)\,\boldsymbol{\epsilon}_H,\quad
  \tilde{\boldsymbol{X}}_P^{k} = \boldsymbol{X}_P^{k} + (1-\rho)\,\boldsymbol{\epsilon}_P,
\end{equation}
where $\boldsymbol{\epsilon}_H, \boldsymbol{\epsilon}_P \sim \mathcal{N}(\boldsymbol{0}, \sigma_{\epsilon}^2 \mathbf{I})$ denote independent Gaussian noise. These perturbations simulate imperfect retrieval and weak-support scenarios that are more likely to occur under distribution shift, where retrieval similarity is generally lower than in training data. By exposing the model to degraded memory conditions during training, REMIND is encouraged to reduce over-reliance on highly confident retrievals and improve generalization under challenging shift conditions.

\vspace{-2mm}
\section{Experiments}
\vspace{-2mm}
\subsection{Experimental Settings}
\vspace{-2mm}
\paragraph{Datasets and Metrics.}
We conduct experiments on three real-world spatiotemporal datasets, namely CAMELS~\cite{addor2017camels}, PEMS~\cite{chen2021trafficstream}, and ENERGY~\cite{zhou2022sdwpf}, which cover hydrology, transportation, and energy domains. These datasets span diverse temporal and spatial resolutions and cover long observation periods, leading to pronounced distribution shifts, as shown in Fig.~\ref{fig:data_distribution}. We evaluate REMIND using both deterministic forecasting metrics, including MAE, RMSE, and MAPE, and probabilistic metrics, including CRPS~\cite{matheson1976scoring} and the Energy Score (ES)~\cite{an2024spatio}. (Details in App.~\ref{app:dataset} and~\ref{app:metrics})
\vspace{-3mm}
\paragraph{Baselines.}
We compare REMIND with eight advanced probabilistic forecasting methods across the two categories defined above. The parametric distribution models include DeepAR~\cite{salinas2020deepar}, LODE~\cite{rubanova2019latent}, DER~\cite{amini2020deep}, and SDER~\cite{wu2024evidence}. The generative distribution models include PriSTI~\cite{liu2023pristi}, DiffSTG~\cite{wen2023diffstg}, STGNF~\cite{an2024spatio}, and STMPN~\cite{an2025spatio}. In addition, we compare REMIND with distribution shift methods, including ShifTS~\cite{zhao2025tackling}, STONE~\cite{wang2024stone}, TFPS~\cite{sun2024learning}, and STRAP~\cite{zhang2025strap}. As these methods are deterministic and do not model predictive uncertainty, the comparison is conducted using deterministic metrics.
\vspace{-3mm}
\paragraph{Implementation Details.}
All datasets are split into training, validation, and test sets with a ratio of 6:2:2, and Z-score normalization is applied using training statistics. The model predicts the next 12 time steps based on the previous 12 observations. We adopt the Adam optimizer~\cite{kingma2014adam} with an initial learning rate of $1\mathrm{e}{-4}$. For fair comparison, baseline hyperparameters are carefully tuned around their recommended or default settings. All results are averaged over three runs with different random seeds. Additional implementation details settings are provided in App.~\ref{app:implementation}.

\vspace{-2mm}
\subsection{Performance Comparison}

\setlength{\tabcolsep}{5.2pt}
\renewcommand{\arraystretch}{1.0}
\begin{table*}[th]
\footnotesize
\centering
\vspace{-0mm}
\caption{Overall performance comparison. The best is \textbf{bold} and the second best is \underline{underlined}.}
\vspace{0mm}
\label{tab:prob}
\begin{tabular}{c|c|cccc|cccc|c}
\toprule
Dataset & Metric
& DeepAR & LODE & DER & SDER & PriSTI & DiffSTG & STGNF & STMPN & \textbf{REMIND} \\
\midrule

\multirow{5}{*}{CAMELS}
 & MAE  & 9.85 & 10.52 & 9.68 & 9.32 & 9.05 & \underline{8.93} & 8.96 & 9.23 & \textbf{8.73} \\
 & RMSE & 38.95 & 34.87 & 38.73 & 37.95 & 36.84 & \underline{34.12} & 35.93 & 35.93 & \textbf{32.26} \\
 & MAPE & 29.53 & 45.14 & 37.85 & 38.24 & 35.65 & 30.96 & \underline{29.47} & 34.25 & \textbf{25.30} \\
 & CRPS & 0.3611 & 0.3252 & 0.4210 & 0.3056 & 0.2955 & 0.3028 & 0.2818 & \underline{0.2577} & \textbf{0.2232} \\
 & ES   & 28.73 & 31.27 & 36.99 & 27.67 & 28.18 & 27.46 & \underline{26.46} & 29.86 & \textbf{26.04} \\

\midrule

\multirow{5}{*}{PEMS}
 & MAE  & 19.82 & 21.45 & 20.25 & 19.64 & 19.38 & 19.13 & 18.55 & \underline{17.91} & \textbf{16.63} \\
 & RMSE & 32.95 & 34.82 & 33.42 & 32.64 & 32.10 & 31.75 & 30.52 & \underline{29.28} & \textbf{26.88} \\
 & MAPE & 20.85 & 23.93 & 22.10 & 21.66 & 20.31 & 20.98 & 19.46 & \underline{18.60} & \textbf{17.76} \\
 & CRPS & 0.0942 & 0.0985 & 0.1013 & 0.0915 & 0.0906 & 0.0898 & 0.0887 & \underline{0.0868} & \textbf{0.0854} \\
 & ES   & 520.75 & 560.48 & 548.93 & 505.37 & 502.85 & 498.69 & \underline{472.92} & 486.53 & \textbf{453.21} \\

\midrule

\multirow{5}{*}{ENERGY}
 & MAE  & 2.12 & 2.45 & 2.25 & 2.18 & 2.05 & 2.08 & \underline{1.93} & 2.30 & \textbf{1.72} \\
 & RMSE & 3.95 & 3.62 & 3.98 & 3.85 & 3.23 & 3.88 & \underline{2.95} & 4.77 & \textbf{2.39} \\
 & MAPE & 24.84 & 28.97 & 27.60 & 26.95 & 25.41 & 24.74 & \underline{23.17} & 25.51 & \textbf{21.49} \\
 & CRPS & 0.1628 & 0.1452 & 0.1745 & 0.1397 & 0.1345 & \underline{0.1218} & 0.1372 & 0.1437 & \textbf{0.1027} \\
 & ES   & 19.66 & 20.85 & 21.48 & 19.12 & 18.97 & \underline{18.10} & 19.23 & 20.72 & \textbf{16.92} \\

\bottomrule
\end{tabular}
\end{table*}

\begin{table}[t]
  \centering
  \vspace{-3mm}
  \begin{minipage}{0.48\linewidth}
    \centering
    \caption{Ablation study on core architecture.}
    \vspace{0mm}
    \label{tab:ablation_core_energy}
    \setlength{\tabcolsep}{5pt}

\footnotesize

\vspace{0mm}

\centering

\begin{tabular}{l|c c c c c}

\toprule

Variant & MAE & RMSE & MAPE & CRPS & ES \\

\midrule

w/o Mem   & 2.90 & 4.03 & 46.29 & 0.1664 & 27.25 \\

w/o Dev   & 3.12 & 4.14 & 46.75 & 0.1935 & 30.43 \\

w/o Gate  & 3.01 & 4.09 & 48.48 & 0.1932 & 30.67 \\

\textbf{REMIND} & \textbf{1.72} & \textbf{2.39} & \textbf{21.49} & \textbf{0.1027} & \textbf{16.92} \\

\bottomrule

\end{tabular}

\vspace{-5mm}
  \end{minipage}
  \hfill
  \begin{minipage}{0.48\linewidth}
    \centering
    \caption{Ablation study on key mechanisms.}
    \vspace{0mm}
    \label{tab:ablation_key_energy}
    \setlength{\tabcolsep}{5pt}
\footnotesize
\vspace{0mm}

\centering
\begin{tabular}{l|c c c c c}
\toprule
Variant & MAE & RMSE & MAPE & CRPS & ES \\
\midrule
w/o MD   & 2.06 & 2.85 & 30.38 & 0.1268 & 19.63 \\
w/o MP   & 1.78 & 2.47 & 23.28 & 0.1050 & 17.29 \\
w/o GM   & 1.74 & 2.41 & 21.73 & 0.1173 & 19.57 \\
\textbf{REMIND} & \textbf{1.72} & \textbf{2.39} & \textbf{21.49} & \textbf{0.1027} & \textbf{16.92} \\
\bottomrule
\end{tabular}
\vspace{-5mm}
  \end{minipage}
  \vspace{-0mm}
  \end{table}

\begin{figure*}[t]
  \vspace{-0mm}
\centering
\begin{minipage}[]{0.38\linewidth}
  \centering
    \includegraphics[width=1\linewidth]{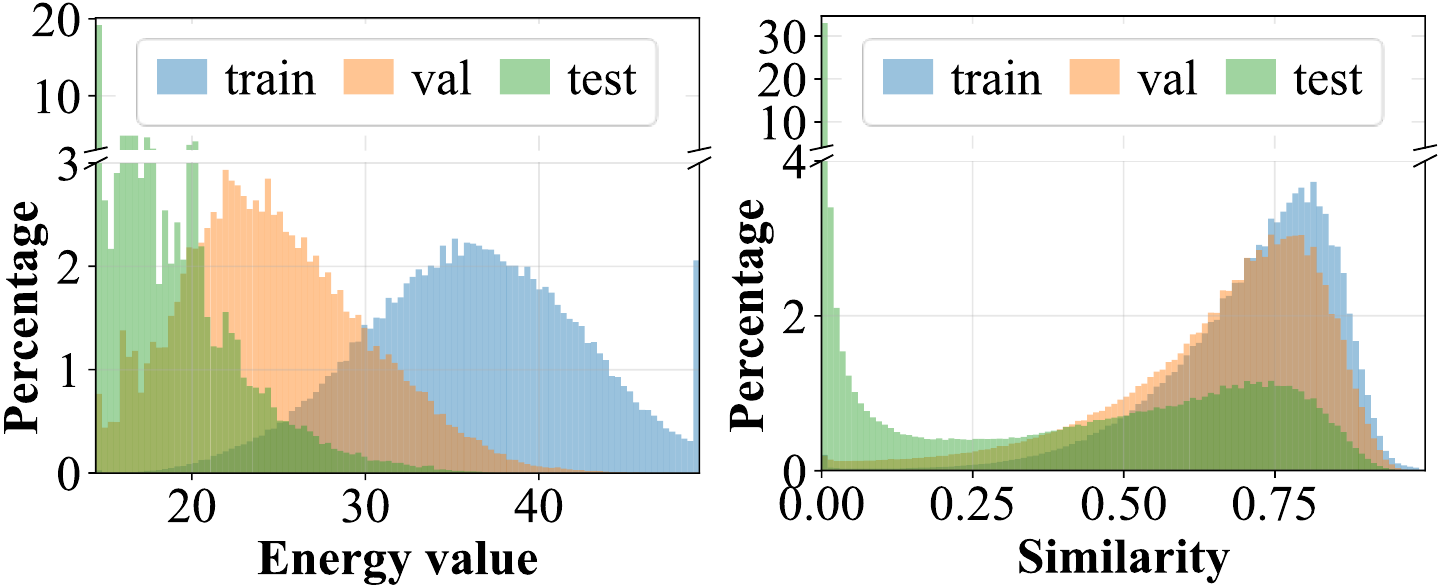}
    \vspace{-5mm}
    \caption{Distribution on ENERGY.}
    \label{fig:data_distribution}
    
\end{minipage}
\hfill
\begin{minipage}[]{0.6\linewidth}
  \centering
  \captionof{table}{Comparison with distribution shift methods.}
  \label{tab:shift_camels}
  \resizebox{\linewidth}{!}{%




\setlength{\tabcolsep}{5pt}
\renewcommand{\arraystretch}{1}
\footnotesize
\centering
\begin{tabular}{c|ccccc}
\toprule
Metric & ShiftTS & STONE & TFPS & STRAP & \textbf{REMIND} \\
\midrule

MAE  & 9.51\std{0.18} & \underline{8.85}\std{0.16} & 9.16\std{0.17} & 8.95\std{0.15} & \textbf{8.73}\std{0.14} \\
RMSE & 34.67\std{0.62} & \underline{33.81}\std{0.58} & 34.35\std{0.60} & 34.33\std{0.57} & \textbf{32.26}\std{0.55} \\
MAPE & 27.24\std{0.85} & 28.05\std{0.92} & \underline{27.23}\std{0.88} & 29.85\std{1.05} & \textbf{25.30}\std{0.80} \\
\bottomrule
\end{tabular}%
  }
\end{minipage}
\vspace{-5mm}
\end{figure*}



As shown in Tab.~\ref{tab:prob}, REMIND achieves the best performance across all datasets and evaluation metrics. Compared with conventional parametric models, REMIND improves distributional expressiveness through the hierarchical Gaussian mixture while retaining the efficiency of likelihood-based prediction. Although generative methods provide stronger distribution modeling capacity, they do not estimate historical support. In contrast, REMIND leverages retrieval as shift signals to adaptively balance memory reuse and deviation-aware prediction, leading to better generalization under distribution shift. We further compare REMIND with representative distribution shift methods in Tab.~\ref{tab:shift_camels}. Although these baselines mainly focus on deterministic prediction, REMIND still achieves the best performance on CAMELS. The results demonstrate that characterizing shift degree and adaptively adjusting prediction strategies are more effective than a fixed adaptation strategy to all samples.

\vspace{-2mm}
\subsection{Ablation Study}
\vspace{-2mm}
Tabs.\ref{tab:ablation_core_energy} and\ref{tab:ablation_key_energy} report two groups of ablation studies on ENERGY. Removing the memory-calibrated probabilistic branch (Mem), deviation-aware probabilistic branch (Dev), or shift-adaptive gate (Gate) all leads to clear performance degradation, confirming that REMIND relies on the cooperation of retrieval reuse, deviation-aware prediction, and adaptive fusion. Among these components, removing the Dev causes the largest degradation, highlighting the importance of modeling deviations from historical patterns under strong distribution shift. The gate is also critical, showing that continuously varying shift conditions cannot be handled by static fusion. We further analyze several key mechanisms, including memory dispersion (MD), memory perturbation (MP), and Gaussian mixture modeling (GM). All three contribute to the final performance. In particular, removing memory dispersion causes the largest drop, supporting its role as an indicator of retrieval reliability under distribution shift. Memory perturbation improves robustness by reducing over-reliance on highly confident retrieval during training, while Gaussian mixture modeling improves probabilistic quality by better capturing diverse future modes.

\vspace{-2mm}
\subsection{Hyperparameter Study}
\vspace{-3.5mm}
\begin{figure}[ht]
  \centering
  \vspace{-0mm}
  \includegraphics[width=\linewidth]{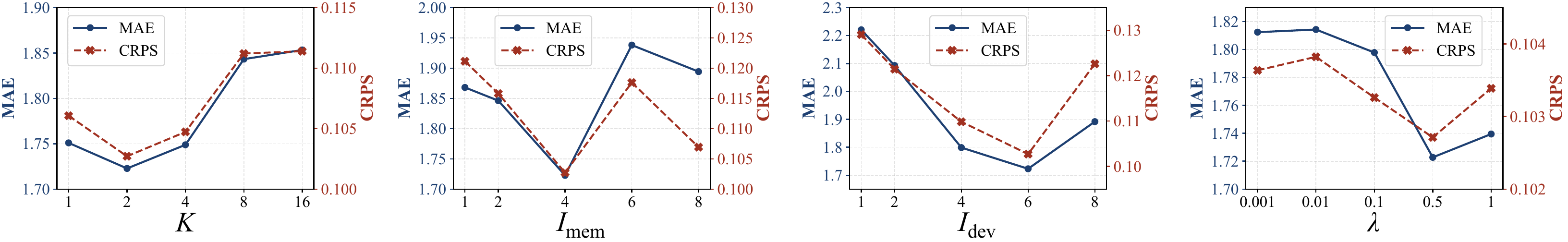}
  \vspace{-6mm}
  \caption{Hyperparameter sensitivity on ENERGY.}
  \label{fig:hyper_energy}
  \vspace{-3.5mm}
\end{figure}

Fig.~\ref{fig:hyper_energy} presents the hyperparameter analysis of REMIND. The results for $I_{\mathrm{mem}}$ and $I_{\mathrm{dev}}$ are consistent with the design of the hierarchical Gaussian mixture: the memory-calibrated probabilistic branch requires fewer components because retrieved targets already provide structured future patterns, while the deviation-aware probabilistic branch benefits from more components to model diverse future evolutions under severe distribution shift. In addition, the hyperparameter analysis of $\lambda$ indicates that the gate benefits from soft shift-guided regularization without being overly constrained.

\vspace{-2mm}
\subsection{Gate Behavior Analysis}
\vspace{-3.5mm}
\begin{figure}[th]
  \centering
  \makebox[\linewidth][c]{%
  \resizebox{1.1\linewidth}{!}{
  \begin{minipage}{0.62\linewidth}
    \centering
    \includegraphics[width=\linewidth]{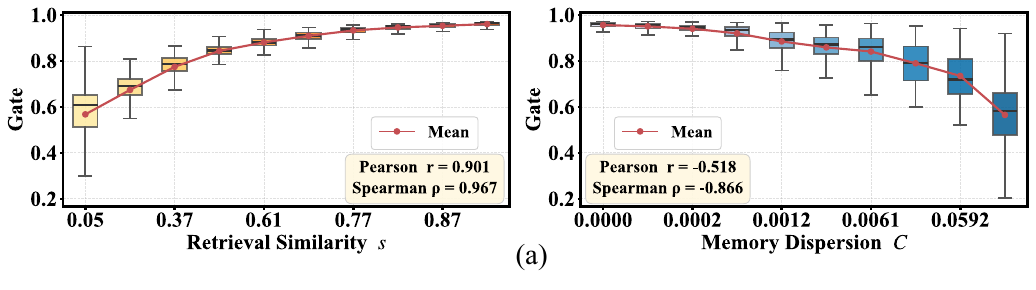}
    \vspace{-5mm}
  \end{minipage}
  \hfill
  \begin{minipage}{0.37\linewidth}
    \centering
    \includegraphics[width=\linewidth]{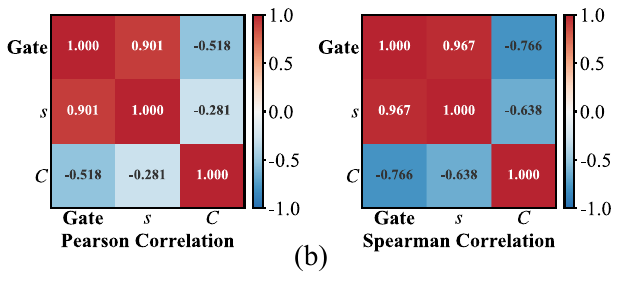}
    \vspace{-4mm}
  \end{minipage}
  }
  }
  \vspace{-3mm}
  \caption{(a) Box plots of learned gate values across retrieval similarity and memory dispersion levels. (b) Pearson and Spearman correlations among learned gate values and these two signals.}
  \vspace{-2mm}
  \label{gate_behavior}
\end{figure}

\begin{figure}[!ht]
  \centering
  \vspace{-3mm}
  \includegraphics[width=\linewidth]{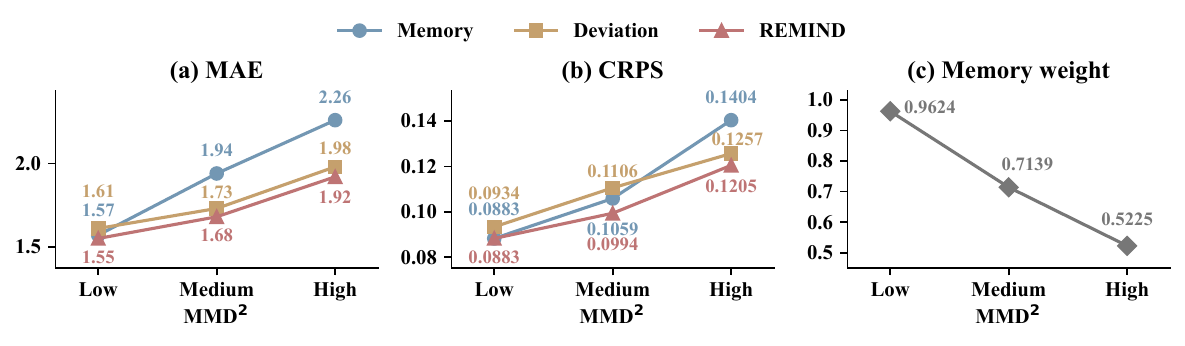}
  \vspace{-6mm}
  \caption{Branch performance and memory reliance across $\text{MMD}^2$ groups on ENERGY.}
  \vspace{-2mm}
  \label{fig:mmd-branches}
  \end{figure}

\begin{table}[t]
\centering
\caption{Memory dispersion analysis on ENERGY. We select the top third of test samples by retrieval similarity and divide them into three groups based on the tertiles of $\boldsymbol{C}$.}
\label{tab:similar-history}
\renewcommand{\arraystretch}{0.82}
\setlength{\aboverulesep}{0.25ex}
\setlength{\belowrulesep}{0.25ex}
\setlength{\tabcolsep}{7pt}
\begin{tabular}{lcccc}
\toprule
& Memory branch & Deviation branch & Mean $\boldsymbol{g}$ & REMIND \\
Dispersion & MAE / CRPS & MAE / CRPS & & MAE / CRPS \\
\midrule
Low & 1.38 / 0.0795 & 1.52 / 0.0860 & 0.91 & 1.35 / 0.0780 \\
Medium & 1.60 / 0.0925 & 1.57 / 0.0910 & 0.66 & 1.52 / 0.0875 \\
High & 1.92 / 0.1110 & 1.83 / 0.1045 & 0.38 & 1.78 / 0.0994 \\
\bottomrule
\end{tabular}
\vspace{-3mm}
\end{table}

We analyze whether the gate responds to retrieval similarity and memory dispersion, and adjusts branch contributions accordingly under distribution shift.

Fig.~\ref{gate_behavior} shows that the gate generally assigns greater memory weight at higher retrieval similarity and lower memory dispersion. This relationship is not strictly monotonic, especially for dispersion, consistent with flexible weighting that combines both signals under soft regularization.

Fig.~\ref{fig:mmd-branches} groups ENERGY test samples into tertiles by squared maximum mean discrepancy (MMD$^2$), computed for each sample against the training data using a Gaussian kernel. Higher MMD$^2$ indicates greater distribution shift. The memory branch performs better under milder shifts, whereas the deviation branch performs better under stronger shifts, accompanied by decreasing memory weights. Fusion achieves the lowest MAE in all groups and improves or matches both branches in CRPS.

Table~\ref{tab:similar-history} further examines samples with high retrieval similarity. As dispersion increases, the better-performing branch shifts from memory to deviation, and the gate accordingly reduces memory weight, demonstrating the role of memory dispersion in assessing retrieval reliability beyond similarity alone. Fusion consistently outperforms both branches, highlighting their complementary strengths and the effectiveness of adaptive fusion.

\vspace{-3mm}
\section{Conclusion}
\vspace{-3.5mm}
We propose REMIND, a retrieval-guided probabilistic framework for spatiotemporal forecasting. By leveraging retrieval as shift degree signals, REMIND adaptively balances memory reuse and deviation aware prediction, improving the generalization of probabilistic forecasting under distribution shift.
\vspace{-0mm}

\vspace{-2mm}
\section*{Acknowledgments}
\vspace{-2mm}
This work was supported by the National Natural Science Foundation of China (No. 62372031).

\clearpage
\bibliography{ref}
\bibliographystyle{unsrt}


\clearpage
\appendix
 
\section{Derivation of the Hierarchical Gaussian Mixture}
\label{app:hgmm}

In this appendix, we provide a detailed derivation of the hierarchical Gaussian mixture
introduced in the Distribution Aware Fusion section of the main paper. We first show that
the gate-weighted combination of the two branch-wise mixtures is itself a flat Gaussian
mixture, then derive its predictive mean and variance, and finally write out the explicit
form of the negative log-likelihood used as the training objective.

\paragraph{Notational conventions.}
The trust gate $\boldsymbol{g} \in [0,1]^{T \times N \times F}$ used in the gate-weighted
fusion is applied element-wise, as are all branch-level parameters
$(\boldsymbol{\mu}_{\mathrm{mem}}, \boldsymbol{\sigma}_{\mathrm{mem}})$,
$(\boldsymbol{\mu}_{\mathrm{dev}}, \boldsymbol{\sigma}_{\mathrm{dev}})$, and the
per-component residuals $(\Delta\boldsymbol{\mu}_i, \Delta\boldsymbol{\sigma}_i)$.
Consequently, the predictive distribution
$p(\boldsymbol{X}_P \mid \boldsymbol{X}_H)$ factorizes element-wise across spatial,
temporal, and feature indices, and every quantity below is to be read as one such scalar
entry. With this convention, we suppress the conditioning input $\boldsymbol{X}_H$ and the
indices $t,n,f$ throughout the derivation. The multivariate spatiotemporal case follows
directly by stacking the per-element results. Two notational points are worth emphasizing
to avoid clashes with the main paper. First, $\boldsymbol{\sigma}_i^{\mathrm{mem}} =
\boldsymbol{\sigma}_{\mathrm{mem}} + \Delta\boldsymbol{\sigma}_i^{\mathrm{mem}}$ denotes a
\emph{component-level standard deviation}. Second, we use
$\bar{\boldsymbol{\sigma}}_{\mathrm{mem}}^2$ to denote the
\emph{branch-level predictive variance} of $p_{\mathrm{mem}}$. These two quantities are different
objects and should not be conflated.

\subsection{From a Hierarchical Mixture to a Flat Mixture}
\label{app:hgmm:flatten}

Recall from the main paper that the memory-calibrated branch and the deviation-aware
branch are parameterized as Gaussian mixtures with $I_{\mathrm{mem}}$ and $I_{\mathrm{dev}}$
components, respectively:
\begin{align}
    p_{\mathrm{mem}}(\boldsymbol{X}_P \mid \boldsymbol{X}_H)
    &= \sum_{i=1}^{I_{\mathrm{mem}}} \boldsymbol{\pi}_i^{\mathrm{mem}} \,
       \mathcal{N}\!\left(\boldsymbol{X}_P \,\big|\, \boldsymbol{\mu}_i^{\mathrm{mem}}, (\boldsymbol{\sigma}_i^{\mathrm{mem}})^2 \right),
       \label{eq:app-pmem} \\
    p_{\mathrm{dev}}(\boldsymbol{X}_P \mid \boldsymbol{X}_H)
    &= \sum_{j=1}^{I_{\mathrm{dev}}} \boldsymbol{\pi}_j^{\mathrm{dev}} \,
       \mathcal{N}\!\left(\boldsymbol{X}_P \,\big|\, \boldsymbol{\mu}_j^{\mathrm{dev}}, (\boldsymbol{\sigma}_j^{\mathrm{dev}})^2 \right),
       \label{eq:app-pdev}
\end{align}
where, consistent with the main paper,
\begin{equation}
    \boldsymbol{\mu}_i^{\mathrm{mem}} = \boldsymbol{\mu}_{\mathrm{mem}} + \Delta\boldsymbol{\mu}_i^{\mathrm{mem}}, \quad
    \boldsymbol{\sigma}_i^{\mathrm{mem}} = \boldsymbol{\sigma}_{\mathrm{mem}} + \Delta\boldsymbol{\sigma}_i^{\mathrm{mem}},
\end{equation}
\begin{equation}
    \boldsymbol{\mu}_j^{\mathrm{dev}} = \boldsymbol{\mu}_{\mathrm{dev}} + \Delta\boldsymbol{\mu}_j^{\mathrm{dev}}, \quad
    \boldsymbol{\sigma}_j^{\mathrm{dev}} = \boldsymbol{\sigma}_{\mathrm{dev}} + \Delta\boldsymbol{\sigma}_j^{\mathrm{dev}},
\end{equation}
and the component weights satisfy
$\sum_{i=1}^{I_{\mathrm{mem}}} \boldsymbol{\pi}_i^{\mathrm{mem}} = 1$ and
$\sum_{j=1}^{I_{\mathrm{dev}}} \boldsymbol{\pi}_j^{\mathrm{dev}} = 1$.

Substituting Eq.~\eqref{eq:app-pmem} and Eq.~\eqref{eq:app-pdev} into the gate-weighted fusion
$p(\boldsymbol{X}_P \mid \boldsymbol{X}_H) = \boldsymbol{g}\, p_{\mathrm{mem}} +
(1-\boldsymbol{g})\, p_{\mathrm{dev}}$ defined in the main paper yields
\begin{equation}
\begin{aligned}
    p(\boldsymbol{X}_P \mid \boldsymbol{X}_H)
    &= \boldsymbol{g} \, p_{\mathrm{mem}}(\boldsymbol{X}_P \mid \boldsymbol{X}_H)
     + (1-\boldsymbol{g}) \, p_{\mathrm{dev}}(\boldsymbol{X}_P \mid \boldsymbol{X}_H) \\
    &= \sum_{i=1}^{I_{\mathrm{mem}}} \boldsymbol{g} \, \boldsymbol{\pi}_i^{\mathrm{mem}} \,
       \mathcal{N}\!\left(\boldsymbol{X}_P \,\big|\, \boldsymbol{\mu}_i^{\mathrm{mem}}, (\boldsymbol{\sigma}_i^{\mathrm{mem}})^2 \right) \\
    &\quad + \sum_{j=1}^{I_{\mathrm{dev}}} (1-\boldsymbol{g}) \, \boldsymbol{\pi}_j^{\mathrm{dev}} \,
       \mathcal{N}\!\left(\boldsymbol{X}_P \,\big|\, \boldsymbol{\mu}_j^{\mathrm{dev}}, (\boldsymbol{\sigma}_j^{\mathrm{dev}})^2 \right).
\end{aligned}
\label{eq:app-flat}
\end{equation}
Eq.~\eqref{eq:app-flat} shows that the hierarchical mixture is equivalent to a flat
Gaussian mixture with $I_{\mathrm{mem}} + I_{\mathrm{dev}}$ components, whose effective
mixing weights are
\begin{equation}
    \tilde{\boldsymbol{\pi}}_i = \boldsymbol{g} \, \boldsymbol{\pi}_i^{\mathrm{mem}} \;\; (i=1,\dots,I_{\mathrm{mem}}), \qquad
    \tilde{\boldsymbol{\pi}}_{I_{\mathrm{mem}}+j} = (1-\boldsymbol{g}) \, \boldsymbol{\pi}_j^{\mathrm{dev}} \;\; (j=1,\dots,I_{\mathrm{dev}}).
    \label{eq:app-flatweights}
\end{equation}
The effective weights remain a valid probability distribution, since
\begin{equation}
    \sum_{i=1}^{I_{\mathrm{mem}}} \boldsymbol{g} \, \boldsymbol{\pi}_i^{\mathrm{mem}}
    + \sum_{j=1}^{I_{\mathrm{dev}}} (1-\boldsymbol{g}) \, \boldsymbol{\pi}_j^{\mathrm{dev}}
    = \boldsymbol{g} + (1-\boldsymbol{g}) = 1.
\end{equation}
This flattening view is convenient for analysis and for evaluating the log-likelihood, while
the hierarchical view of the gate-weighted fusion is more interpretable. The inner level
captures intra-branch multimodality, and the outer level captures distribution-aware
adaptation between the two branches.

\subsection{Predictive Mean}
\label{app:hgmm:mean}

We first compute the \emph{branch-level} predictive mean of each branch from its own
mixture in Eq.~\eqref{eq:app-pmem} and Eq.~\eqref{eq:app-pdev}. To distinguish these from the
component-level means $\boldsymbol{\mu}_i^{\mathrm{mem}}, \boldsymbol{\mu}_j^{\mathrm{dev}}$,
we denote them with a bar:
\begin{equation}
    \bar{\boldsymbol{\mu}}_{\mathrm{mem}} \;\triangleq\; \mathbb{E}_{p_{\mathrm{mem}}}[\boldsymbol{X}_P]
    = \sum_{i=1}^{I_{\mathrm{mem}}} \boldsymbol{\pi}_i^{\mathrm{mem}} \, \boldsymbol{\mu}_i^{\mathrm{mem}}, \qquad
    \bar{\boldsymbol{\mu}}_{\mathrm{dev}} \;\triangleq\; \mathbb{E}_{p_{\mathrm{dev}}}[\boldsymbol{X}_P]
    = \sum_{j=1}^{I_{\mathrm{dev}}} \boldsymbol{\pi}_j^{\mathrm{dev}} \, \boldsymbol{\mu}_j^{\mathrm{dev}}.
    \label{eq:app-branchmean}
\end{equation}
Taking the expectation over the flat mixture in Eq.~\eqref{eq:app-flat} and using linearity,
the overall predictive mean becomes a gate-weighted combination of the two branch means:
\begin{equation}
    \mathbb{E}[\boldsymbol{X}_P]
    = \boldsymbol{g} \, \bar{\boldsymbol{\mu}}_{\mathrm{mem}} + (1-\boldsymbol{g}) \, \bar{\boldsymbol{\mu}}_{\mathrm{dev}}.
    \label{eq:app-mean}
\end{equation}

\subsection{Predictive Variance}
\label{app:hgmm:variance}

For the variance, we again work at the branch level. We define the \emph{branch-level
predictive variance} as $\bar{\boldsymbol{\sigma}}_{\mathrm{mem}}^2 \triangleq
\mathrm{Var}_{p_{\mathrm{mem}}}[\boldsymbol{X}_P]$ and analogously for the deviation
branch. This is an aggregate quantity over all $I_{\mathrm{mem}}$ components and should
not be confused with any single $(\boldsymbol{\sigma}_i^{\mathrm{mem}})^2$. Using the
standard identity $\mathrm{Var}[Z] = \mathbb{E}[Z^2] - (\mathbb{E}[Z])^2$ for a mixture, we
obtain
\begin{align}
    \bar{\boldsymbol{\sigma}}_{\mathrm{mem}}^2
    &= \sum_{i=1}^{I_{\mathrm{mem}}} \boldsymbol{\pi}_i^{\mathrm{mem}} \!\left[(\boldsymbol{\sigma}_i^{\mathrm{mem}})^2 + (\boldsymbol{\mu}_i^{\mathrm{mem}})^2 \right] - \bar{\boldsymbol{\mu}}_{\mathrm{mem}}^2,
       \label{eq:app-memvar} \\
    \bar{\boldsymbol{\sigma}}_{\mathrm{dev}}^2
    &= \sum_{j=1}^{I_{\mathrm{dev}}} \boldsymbol{\pi}_j^{\mathrm{dev}} \!\left[(\boldsymbol{\sigma}_j^{\mathrm{dev}})^2 + (\boldsymbol{\mu}_j^{\mathrm{dev}})^2 \right] - \bar{\boldsymbol{\mu}}_{\mathrm{dev}}^2.
       \label{eq:app-devvar}
\end{align}

Applying the same identity to the flat mixture in Eq.~\eqref{eq:app-flat} gives
\begin{equation}
    \mathrm{Var}[\boldsymbol{X}_P]
    = \boldsymbol{g} \!\left(\bar{\boldsymbol{\sigma}}_{\mathrm{mem}}^2 + \bar{\boldsymbol{\mu}}_{\mathrm{mem}}^2\right)
    + (1-\boldsymbol{g}) \!\left(\bar{\boldsymbol{\sigma}}_{\mathrm{dev}}^2 + \bar{\boldsymbol{\mu}}_{\mathrm{dev}}^2\right)
    - \big( \boldsymbol{g} \, \bar{\boldsymbol{\mu}}_{\mathrm{mem}} + (1-\boldsymbol{g}) \, \bar{\boldsymbol{\mu}}_{\mathrm{dev}} \big)^2.
\end{equation}
Expanding the square and collecting terms yields the following more interpretable
decomposition:
\begin{equation}
    \mathrm{Var}[\boldsymbol{X}_P]
    = \boldsymbol{g} \, \bar{\boldsymbol{\sigma}}_{\mathrm{mem}}^2
    + (1-\boldsymbol{g}) \, \bar{\boldsymbol{\sigma}}_{\mathrm{dev}}^2
    + \boldsymbol{g}(1-\boldsymbol{g}) \, \big(\bar{\boldsymbol{\mu}}_{\mathrm{mem}} - \bar{\boldsymbol{\mu}}_{\mathrm{dev}}\big)^{2}.
    \label{eq:app-vardecomp}
\end{equation}
Eq.~\eqref{eq:app-vardecomp} decomposes the predictive uncertainty into two sources.
The first term is a gate-weighted average of the within-branch mixture variances and
reflects how uncertain each branch is on its own. The second term captures the discrepancy
between the memory-calibrated and deviation-aware branch means. It increases when the two
branches disagree and reaches its largest value when the gate is most ambivalent,
namely when $\boldsymbol{g}\to\tfrac{1}{2}$. As a result, the model reports higher
uncertainty in the transitional regime between in-distribution and out-of-distribution
inputs even when each branch is individually confident.

\subsection{Explicit Form of the Negative Log-Likelihood}
\label{app:hgmm:nll}

We now derive the explicit form of the negative log-likelihood referenced in the Training
Objective section of the main paper namely the unified mixture form introduced after the
definition of $\mathcal{L}_{\mathrm{NLL}}$. Substituting Eq.~\eqref{eq:app-flat} directly
into $\mathcal{L}_{\mathrm{NLL}} = -\,\mathbb{E}[\log p(\boldsymbol{X}_P \mid \boldsymbol{X}_H)]$
gives
\begin{equation}
\begin{aligned}
    \mathcal{L}_{\mathrm{NLL}}
    = -\,\mathbb{E}\!\left[\log \Bigg( \right.&
        \sum_{i=1}^{I_{\mathrm{mem}}} \boldsymbol{g} \, \boldsymbol{\pi}_i^{\mathrm{mem}} \,
        \mathcal{N}\!\left(\boldsymbol{X}_P \,\big|\, \boldsymbol{\mu}_i^{\mathrm{mem}}, (\boldsymbol{\sigma}_i^{\mathrm{mem}})^2 \right) \\
    & \left. + \sum_{j=1}^{I_{\mathrm{dev}}} (1-\boldsymbol{g}) \, \boldsymbol{\pi}_j^{\mathrm{dev}} \,
        \mathcal{N}\!\left(\boldsymbol{X}_P \,\big|\, \boldsymbol{\mu}_j^{\mathrm{dev}}, (\boldsymbol{\sigma}_j^{\mathrm{dev}})^2 \right) \Bigg) \right].
\end{aligned}
\label{eq:app-nll}
\end{equation}
For numerical stability, we evaluate Eq.~\eqref{eq:app-nll} with the log-sum-exp trick. Let
$\tilde{\boldsymbol{\pi}}_k$ and $\big(\tilde{\boldsymbol{\mu}}_k, \tilde{\boldsymbol{\sigma}}_k^2\big)$
denote the effective weight and parameters of the $k$-th component of the flat mixture,
where $k=1,\dots,I_{\mathrm{mem}}+I_{\mathrm{dev}}$. These quantities are defined in
Eq.~\eqref{eq:app-flatweights}. Then
\begin{equation}
    \mathcal{L}_{\mathrm{NLL}}
    = -\,\mathbb{E}\!\left[
        \operatorname{LSE}_{k}\!\left(
            \log \tilde{\boldsymbol{\pi}}_k
            - \tfrac{1}{2}\log\!\big(2\pi\tilde{\boldsymbol{\sigma}}_k^2\big)
            - \tfrac{(\boldsymbol{X}_P - \tilde{\boldsymbol{\mu}}_k)^2}{2\tilde{\boldsymbol{\sigma}}_k^2}
        \right)
    \right],
\end{equation}
where $\operatorname{LSE}_k(\cdot) \triangleq \log\!\sum_k \exp(\cdot)$. Combining this with
the auxiliary similarity- and dispersion-aware regularizers introduced in the main paper
gives the full training objective $\mathcal{L} = \mathcal{L}_{\mathrm{NLL}} + \lambda
(\mathcal{L}_{\mathrm{sim}} + \mathcal{L}_{\mathrm{dis}})$. Setting
$\boldsymbol{g}\in\{0,1\}$ in Eq.~\eqref{eq:app-mean} and Eq.~\eqref{eq:app-vardecomp}
reduces the model to a single-branch mixture, and setting
$I_{\mathrm{mem}}=I_{\mathrm{dev}}=1$ reduces it to a single-Gaussian forecaster.

\section{Discussion of Spatiotemporal Retrieval}
\label{app:retrieval}
\begin{figure}[ht]
    \centering
    \makebox[\linewidth][c]{%
    \includegraphics[width=1.0\linewidth]{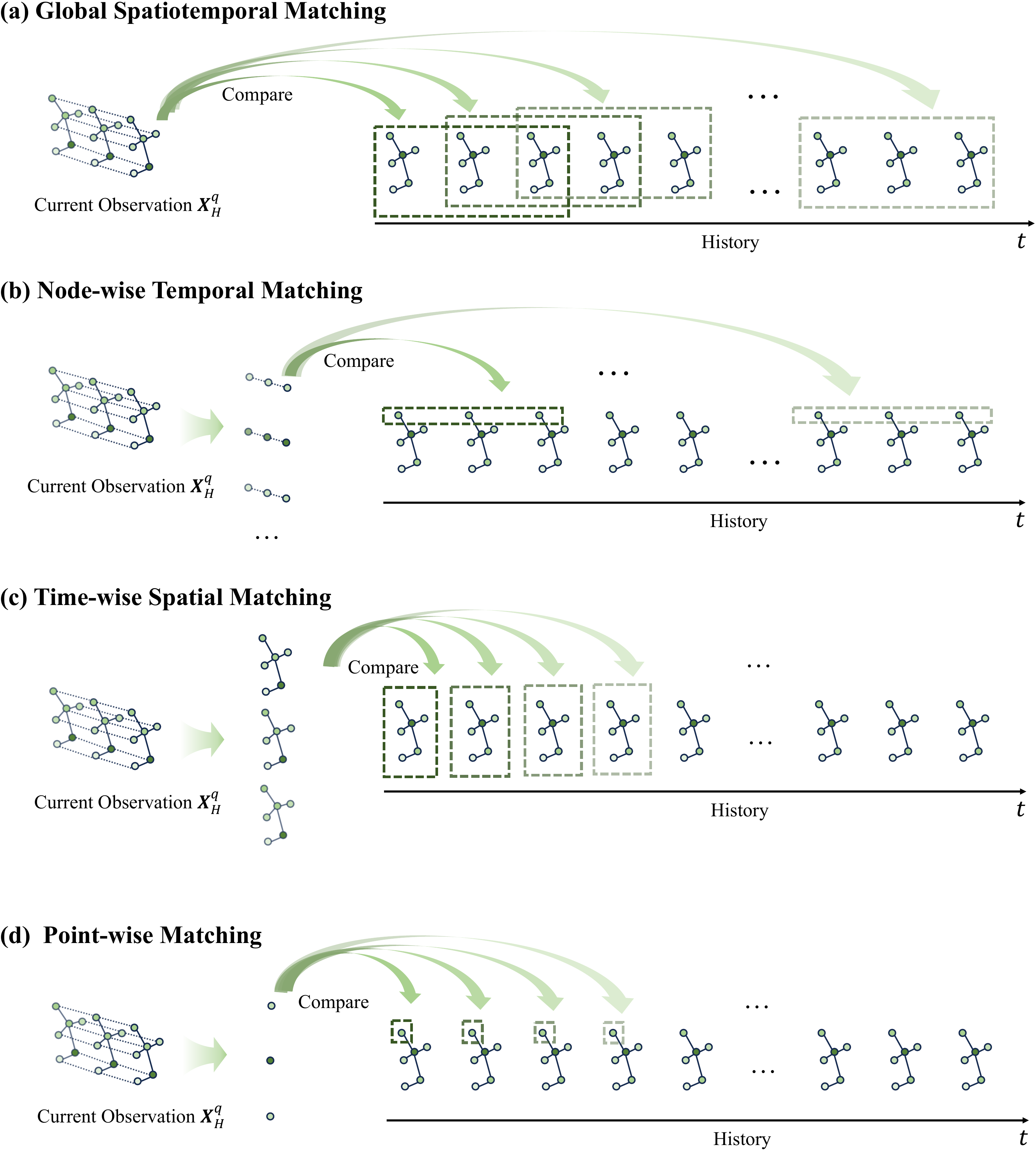}
    }
    \caption{Four levels of spatiotemporal retrieval.}
    \label{app:retrieval_fig}
    \vspace{0mm}
  \end{figure}

Let $\mathcal{D} = \{(\boldsymbol{X}_H^{i}, \boldsymbol{X}_P^{i})\}_{i=1}^R$ denote a memory bank of $R$ historical observation-target pairs. In an ideal spatiotemporal retrieval mechanism, given a query observation $\boldsymbol{X}_H^{q}$, the system aims to rapidly and efficiently retrieve the top $K$ most similar samples $\mathcal{D}_{\mathrm{ret}} = \{\boldsymbol{X}_H^{k}, \boldsymbol{X}_P^{k}, \boldsymbol{s}_k\}_{k=1}^K$, where $\boldsymbol{s}_k$ denotes the similarity score between $\boldsymbol{X}_H^{q}$ and $\boldsymbol{X}_H^{k}$.

\begin{table}[ht]
    \centering
    \footnotesize
    \setlength{\tabcolsep}{4pt}
    \renewcommand{\arraystretch}{1.1}
    \caption{Comparison of different spatiotemporal retrieval strategies.}
    \label{tab:retrieval_compare}
    \begin{tabular}{c|ccc|ccc|ccc}
    \toprule
    & \multicolumn{3}{c|}{CAMELS} 
    & \multicolumn{3}{c|}{ENERGY} 
    & \multicolumn{3}{c}{PEMS} \\
    \cline{2-10}
    Method 
    & MAE & MAPE & Time 
    & MAE & MAPE & Time 
    & MAE & MAPE & Time \\
    \midrule
    Global 
    & 11.73 & 43.65 & 1.53 
    & 8.85 & 109.76 & 920.53 
    & 27.27 & 43.54 & 11407.35 \\
    
    Node-wise 
    & 10.87 & 36.40 & 3.90 
    & 7.96 & 104.70 & 3569.83 
    & 19.29 & 28.12 & 21179.26 \\
    
    Time-wise 
    & 14.00 & 58.26 & 7.15 
    & 8.76 & 108.58 & 467.50 
    & 27.41 & 44.42 & 4526.34 \\
    
    Point-wise 
    & 16.09 & 60.87 & 8.66 
    & 10.97 & 139.52 & 1239.79 
    & 41.72 & 53.20 & 15527.86 \\
    
    \textbf{Ours} 
    & 10.92 & 37.81 & 1.83 
    & 8.04 & 104.02 & 957.53 
    & 19.02 & 27.93 & 12377.27 \\
    \bottomrule
    \end{tabular}
    \end{table}

Retrieving spatiotemporal data is non-trivial due to the entanglement of spatial and temporal dimensions. Accordingly, spatiotemporal retrieval can be formulated at four levels: global spatiotemporal matching, node-wise temporal matching, time-wise spatial matching, and point-wise matching. Global spatiotemporal matching preserves overall structure with high efficiency but enforces strict temporal alignment across nodes. In contrast, node-wise temporal matching retrieves patterns independently for each node based on its temporal dynamics, enabling more flexible and fine-grained retrieval, but requiring independent matching for each node and thus increasing computational cost. Meanwhile, time-wise spatial matching disrupts temporal continuity, and point-wise matching ignores global spatiotemporal structure, limiting their ability to capture coherent spatiotemporal patterns.

Retrieving spatiotemporal data is challenging due to the strong coupling between spatial and temporal dependencies. Depending on how spatial and temporal structures are matched, spatiotemporal retrieval can be categorized into four levels: global spatiotemporal matching, node-wise temporal matching, time-wise spatial matching, and point-wise matching.

Global spatiotemporal matching treats the entire observation sequence as a unified spatiotemporal pattern and retrieves historical samples according to overall similarity. Given $M$ historical samples and flattened feature dimension $D=T\times N\times F$, its retrieval complexity is approximately $\mathcal{O}(MD)$ for each query sample. This strategy preserves global structure and achieves high retrieval efficiency because matching is performed only once at the sample level. However, it enforces strict temporal alignment across all nodes, making it difficult to capture local temporal misalignment and heterogeneous regional dynamics under distribution shift.

Node-wise temporal matching retrieves historical patterns independently for each node based on temporal dynamics. For each query node, retrieval is performed against node-level candidates, resulting in a complexity of approximately $\mathcal{O}(NMD_t)$, where $D_t=T\times F$ denotes the temporal feature dimension for each node. Compared with global retrieval, this strategy relaxes strict node-level temporal alignment and enables more flexible reuse of local historical patterns, especially when different regions evolve asynchronously. This property is particularly important under distribution shift, where different spatial regions may shift at different times, such as localized traffic peaks caused by large events or regional disturbances. However, the independent retrieval process for each node substantially increases both computational and memory costs as the number of nodes grows.

Time-wise spatial matching performs retrieval independently at each time step based on spatial configurations. Although this strategy can model instantaneous spatial similarity, it breaks temporal continuity and may retrieve temporally inconsistent fragments from unrelated periods. For example, two observations with similar instantaneous values may correspond to completely different temporal states, such as rising and declining phases of system dynamics. As a result, coherent temporal evolution patterns are difficult to preserve.

Point-wise matching performs retrieval at the individual spatiotemporal point level. This strategy provides maximum flexibility and the finest retrieval granularity, but completely ignores higher-level spatiotemporal structure. Consequently, retrieved points may lack semantic consistency and fail to represent meaningful system-level dynamics.

Based on these observations, REMIND adopts a coarse-to-fine spatiotemporal retrieval strategy that combines global spatiotemporal retrieval with node-wise temporal retrieval. The first stage efficiently identifies globally relevant historical contexts, while the second stage refines local temporal matching and improves retrieval diversity under heterogeneous distribution shift. In the second stage, retrieval is further allowed across physically connected neighbors and statistically correlated nodes. This design is motivated by the spatial dependency and propagation characteristics of spatiotemporal systems: state transitions occurring at one node may gradually influence neighboring regions, allowing the model to capture relevant temporal dynamics from nearby nodes in advance. In addition, cross-node retrieval increases retrieval diversity and reduces over-reliance on exact node-level matches. However, retrieval is not performed across all nodes, since unrestricted matching would substantially increase computational cost and may introduce misleading patterns from spatially unrelated regions with significantly different dynamics.

As shown in Tab.~\ref{tab:retrieval_compare}, different retrieval strategies exhibit clear trade-offs between retrieval quality and computational efficiency. Global retrieval achieves the highest efficiency but suffers from limited flexibility due to strict temporal alignment across nodes. Node-wise retrieval substantially improves forecasting performance by enabling more flexible local temporal matching, but incurs significantly higher computational cost because retrieval must be performed independently for each node. Time-wise and point-wise retrieval both perform worse, confirming that breaking temporal continuity or ignoring higher-level spatiotemporal structure weakens the ability to capture coherent system dynamics. Overall, the proposed coarse-to-fine retrieval strategy achieves the best balance between retrieval quality and efficiency by combining globally consistent retrieval with locally adaptive refinement.

\section{Experimental Supplementary}
\subsection{Datasets}
\label{app:dataset}
\begin{table}[t]
    \centering
    \caption{Detailed statistics of datasets.}
    \label{tab:data_stats_appendix}
    \setlength{\tabcolsep}{4pt}
    \begin{tabular}{lcccccc}
    \toprule
    Dataset & Domain & \#Nodes & \#Steps & Resolution & Time Span & Feature \\
    \midrule
    CAMELS & Hydrology & 7   & 12699 & 1 day  & 1980-01-01 -- 2014-10-07 & Flow \\
    PEMS   & Traffic   & 581 & 106848 & 5 min  & 2011--2017 (07-10 to 08-31) & Flow \\
    ENERGY & Energy    & 103 & 34560  & 10 min & 240 days                & Wind speed \\
    \bottomrule
    \end{tabular}
    \end{table}
CAMELS~\cite{addor2017camels}, PEMS~\cite{chen2021trafficstream}, and ENERGY~\cite{chen2024expand, zhou2022sdwpf} are three real-world spatiotemporal datasets from hydrology, transportation, and energy domains, respectively. 
As shown Tab.~\ref{tab:data_stats_appendix}, CAMELS contains daily streamflow observations from 7 basins over a long time span from 1980 to 2014, capturing long-term hydrological variability. In this work, we extract data from Basin 13 for evaluation.
PEMS consists of traffic flow data collected at 5-minute intervals over 7 years from 581 sensors, reflecting strong spatiotemporal dynamics. ENERGY includes wind speed measurements at 10-minute resolution over 240 days from 103 locations, representing high-frequency variations in energy systems. As the monitored locations vary over time, we retain only those shared across the entire observation period for both datasets to ensure spatial consistency.

These datasets exhibit diverse temporal and spatial resolutions from minutes to days and long observation horizons, leading to pronounced distribution shifts across time and conditions, which pose significant challenges for reliable probabilistic forecasting.

\subsection{Metrics}
\label{app:metrics}
We evaluate REMIND by standard deterministic metrics, including MAE, RMSE, and MAPE, and uncertainty metrics, including Continuous Ranked Probability Score (CRPS)~\cite{matheson1976scoring} and the Energy Score (ES)~\cite{an2024spatio}.


\textbf{Continuous Ranked Probability Score (CRPS).}
CRPS is a proper scoring rule for evaluating probabilistic forecasts by measuring the discrepancy between the predicted cumulative distribution function and the observed value. It evaluates the entire predictive distribution rather than a single point estimate.
\begin{equation}
    \text{CRPS}(F, x) = \int_{-\infty}^{\infty} \big[F(z) - \mathbb{I}\{x \le z\}\big]^2 dz,
\end{equation}
where, $F(z)$ denotes the predicted cumulative distribution function, $x$ is the ground-truth observation, $\mathbb{I}\{\cdot\}$ is the indicator function.

In practice, CRPS is approximated using a finite set of quantiles $\mathcal{Q}=\{0.05,0.10,\ldots,0.95\}$:
\begin{equation}
    \text{CRPS}(y_{\text{true}}, y_{\text{pred}}) =
\frac{1}{|\mathcal{Q}|}\sum_{q\in\mathcal{Q}}
\frac{L_q(y_{\text{true}}, y_q^{\text{pred}})}{\sum |y_{\text{true}}|},
\end{equation}

\begin{equation}
    L_q(y_{\text{true}}, y_q^{\text{pred}}) =
2\sum \big|(y_q^{\text{pred}}-y_{\text{true}})\cdot(\mathbb{I}[y_{\text{true}}\le y_q^{\text{pred}}]-q)\big|,
\end{equation}
where, $y_{\text{true}}$ is the ground-truth value, $y_{\text{pred}}$ denotes predictive samples, $y_q^{\text{pred}}$ is the predicted $q$-th quantile, $L_q(\cdot)$ is the pinball loss, $|\mathcal{Q}|$ is the number of quantile levels.

Lower CRPS indicates better alignment between the predicted distribution and the ground truth.

\textbf{Energy Score (ES).}
ES is a multivariate proper scoring rule derived from the energy distance, designed for evaluating probabilistic forecasts in high-dimensional and spatiotemporal settings. It simultaneously measures forecast accuracy and predictive dispersion.
\begin{equation}
    \text{ES} =
\mathbb{E}\big[||y_{\text{pred}} - y_{\text{true}}||_2\big] -
\frac{1}{2}\mathbb{E}\big[||y_{\text{pred}}^{(1)} - y_{\text{pred}}^{(2)}||_2\big],
\end{equation}
where, $y_{\text{pred}}$ denotes predictive samples drawn from the forecast distribution, $y_{\text{true}}$ is the ground-truth observation, $||\cdot||_2$ is the $\ell_2$ norm over the spatial dimension, $y_{\text{pred}}^{(1)}$ and $y_{\text{pred}}^{(2)}$ are two independent samples from the predictive distribution.

The first term measures the expected distance between predictions and observations (accuracy), while the second term captures the diversity among predictive samples (uncertainty representation). Lower ES values indicate more accurate and well-calibrated probabilistic forecasts.

\subsection{Implementation Details.}
\label{app:implementation}
All experiments are implemented in PyTorch 2.4.0 and conducted on an NVIDIA GeForce RTX A4000 GPU. For each dataset, baseline hyperparameters are carefully tuned around their recommended or default settings to ensure optimal performance. All results are averaged over three runs with different random seeds. All datasets are split into training, validation, and test sets with a ratio of 6:2:2, and Z-score normalization is applied. The model predicts the next 12 time steps based on the previous 12 observations. We adopt the Adam optimizer~\cite{kingma2014adam} with a batch size of 12 and an initial learning rate of $1\mathrm{e}{-4}$. The hidden dimension is set to 64, and the memory perturbation strength is set to 0.2. The number of retrieved samples $K$, the numbers of Gaussian components $I_{\text{mem}}$ and $I_{\text{dev}}$, and the loss hyperparameter $\lambda$ are determined via hyperparameter tuning.

\subsection{Additional Comparison with Distribution Shift Methods}
\setlength{\tabcolsep}{5pt}
\renewcommand{\arraystretch}{1}
\begin{table}[ht]
\footnotesize
\centering
\caption{Comparison with Distribution Shift Methods on PEMS}
\vspace{0mm}
\label{tab:shift_pems}
\begin{tabular}{c|ccccc}
\toprule
Metric & ShiftTS & STONE & TFPS & STRAP & \textbf{REMIND} \\
\midrule
MAE  & 16.84\std{0.21} & 18.65\std{0.28} & 17.29\std{0.23} & \textbf{16.60}\std{0.20} & \underline{16.63}\std{0.19} \\
RMSE & 27.82\std{0.35} & 30.53\std{0.42} & 30.31\std{0.40} & \textbf{26.32}\std{0.34} & \underline{26.88}\std{0.32} \\
MAPE & 18.00\std{0.48} & 21.21\std{0.65} & \underline{17.69}\std{0.46} & 20.24\std{0.55} & \textbf{17.76}\std{0.44} \\
\bottomrule
\end{tabular}
\end{table}
\setlength{\tabcolsep}{5pt}

\renewcommand{\arraystretch}{1}

\begin{table}[ht]

\footnotesize

\centering

\caption{Comparison with Distribution Shift Methods on ENERGY}

\vspace{0mm}

\label{tab:shift_energy}

\begin{tabular}{c|ccccc}

\toprule

Metric & ShiftTS & STONE & TFPS & STRAP & \textbf{REMIND} \\

\midrule

MAE  & \underline{1.76}\std{0.03} & 2.35\std{0.04} & 1.96\std{0.03} & 2.77\std{0.05} & \textbf{1.72}\std{0.03} \\

RMSE & \underline{2.43}\std{0.04} & 2.93\std{0.05} & 2.49\std{0.04} & 3.75\std{0.06} & \textbf{2.39}\std{0.04} \\

MAPE & 23.82\std{0.55} & \underline{23.51}\std{0.58} & 22.16\std{0.50} & 26.28\std{0.70} & \textbf{21.49}\std{0.48} \\

\bottomrule

\end{tabular}

\end{table}
Tabs.~\ref{tab:shift_pems} and~\ref{tab:shift_energy} provide additional comparisons with deterministic distribution shift methods on PEMS and ENERGY. REMIND remains highly competitive across both datasets, achieving the best or near-best deterministic performance while also preserving probabilistic forecasting capability. On ENERGY, REMIND consistently outperforms the compared shift-aware baselines, showing that its adaptive memory usage is effective under high-frequency dynamics where retrieved patterns can vary substantially over time.

On PEMS, REMIND performs on par with the strongest deterministic baseline and achieves the best result on one of the reported metrics. This is a strong result because the competing methods are specifically optimized for point prediction under distribution shift, whereas REMIND additionally learns predictive uncertainty. The comparison suggests that modeling uncertainty and adapting memory reliance can provide a more stable forecasting strategy under non-stationary traffic patterns.

\subsection{Additional Ablation Study}
\begin{table}[ht]
    \centering
    \begin{minipage}{0.48\linewidth}
      \centering
      \caption{Ablation study on core architecture.}
      \vspace{0mm}
      \label{tab:ablation_core_camels}
      \setlength{\tabcolsep}{5pt}

\footnotesize

\vspace{0mm}

\centering

\begin{tabular}{l|c c c c c}

\toprule

Variant & MAE & RMSE & MAPE & CRPS & ES \\

\midrule

w/o Mem   & 9.51 & 33.19 & 32.04 & 0.2992 & 32.75 \\

w/o Dev   & 9.15 & 32.83 & 28.61 & 0.2309 & 26.79 \\

w/o Gate  & 9.29 & 32.56 & 31.53 & 0.2439 & 29.01 \\

\textbf{REMIND} & \textbf{8.73} & \textbf{32.26} & \textbf{25.30} & \textbf{0.2232} & \textbf{26.04} \\

\bottomrule

\end{tabular}

\vspace{0mm}

    \end{minipage}
    \hfill
    \begin{minipage}{0.48\linewidth}
      \centering
      \caption{Ablation study on key mechanisms.}
      \vspace{0mm}
      \label{tab:ablation_key_camels}
      \setlength{\tabcolsep}{5pt}
\footnotesize
\vspace{0mm}

\centering
\begin{tabular}{l|c c c c c}
\toprule
Variant & MAE & RMSE & MAPE & CRPS & ES \\
\midrule
w/o MD   & 8.93 & 32.30 & 28.04 & 0.2546 & 30.06 \\
w/o MP   & 8.86 & 32.29 & 27.21 & 0.2494 & 29.49 \\
w/o GM   & 8.83 & 32.35 & 26.03 & 0.2405 & 28.20 \\
\textbf{REMIND} & \textbf{8.73} & \textbf{32.26} & \textbf{25.30} & \textbf{0.2232} & \textbf{26.04} \\
\bottomrule
\end{tabular}
\vspace{0mm}
    \end{minipage}
    \end{table}
Tabs.~\ref{tab:ablation_core_camels} and~\ref{tab:ablation_key_camels} report additional ablation results on CAMELS. The trends are consistent with the ENERGY ablations in the main paper: removing any core component weakens performance, confirming that the memory branch, deviation branch, and gate each play a necessary role. The memory branch is especially important for probabilistic quality on CAMELS, where long-term hydrological patterns can provide useful historical support but still require careful calibration. The deviation-aware branch and gate further improve robustness by preventing the model from blindly trusting retrieved futures when the current basin dynamics differ from historical patterns.

The detailed ablations also support the effectiveness of the proposed mechanisms. Memory dispersion(MD) provides an explicit measure of retrieval reliability and improves uncertainty estimation. Memory perturbation(MP) encourages the model to tolerate imperfect memory and reduces sensitivity to retrieval errors. The Gaussian mixture(GM) improves both accuracy and calibration by allowing the predictive distribution to represent multiple plausible hydrological responses. These results show that the gains of REMIND do not come from a single module, but from combining retrieval, shift-aware fusion, and flexible probabilistic modeling.

\subsection{Additional Hyperparameter Study}
\begin{figure}[ht]
    \centering
    \includegraphics[width=\linewidth]{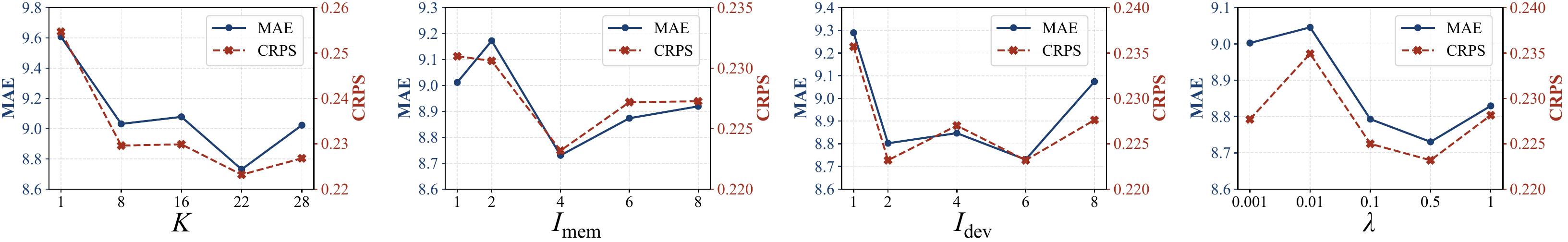}
    \caption{Hyperparameter sensitivity on CAMELS.}
    \label{fig:hyper_camels}
    \end{figure}
Fig.~\ref{fig:hyper_camels} shows the hyperparameter sensitivity on CAMELS. The overall trends are similar to ENERGY. A moderate top-$K$ value performs best because it retrieves enough historical support without introducing excessive irrelevant memory. Too few retrieved samples make the prediction depend on narrow evidence, while too many samples dilute the relevance of the memory bank and increase inconsistency among retrieved futures.

For the Gaussian mixture configuration, the memory-calibrated branch prefers a compact number of components, while the deviation-aware branch benefits from a larger mixture. This again matches the model design: the memory branch already inherits possible modes from retrieved targets, whereas the deviation branch must construct a broader predictive distribution from deviation information. The sensitivity to $\lambda$ also confirms the role of soft gate regularization. A moderate value gives the best balance between likelihood-driven learning and shift-signal guidance; overly weak regularization underuses retrieval statistics, while overly strong regularization restricts the adaptive behavior of the gate.

\subsection{Robustness to Memory Corruption}
\label{sec:memory-corruption}
\begin{table}[!ht]
\centering
\caption{Memory corruption on ENERGY.}
\label{tab:memory-corruption}
\renewcommand{\arraystretch}{0.82}
\setlength{\aboverulesep}{0.25ex}
\setlength{\belowrulesep}{0.25ex}
\setlength{\tabcolsep}{6pt}
\begin{tabular}{cccccccc}
\toprule
& \multicolumn{3}{c}{STRAP} & \multicolumn{3}{c}{REMIND} & \\
\cmidrule(lr){2-4}\cmidrule(lr){5-7}
Noise & MAE & RMSE & MAPE & MAE & RMSE & MAPE & Mean $g$ \\
\midrule
0.0 & 2.77 & 3.75 & 26.28 & 1.72 & 2.39 & 21.49 & 0.74 \\
0.1 & 2.95 & 3.96 & 28.10 & 1.76 & 2.44 & 22.10 & 0.66 \\
0.2 & 3.24 & 4.31 & 31.47 & 1.84 & 2.54 & 23.38 & 0.53 \\
0.4 & 3.78 & 4.93 & 37.85 & 2.02 & 2.76 & 25.91 & 0.36 \\
0.6 & 4.31 & 5.58 & 44.92 & 2.23 & 3.02 & 29.14 & 0.24 \\
\bottomrule
\end{tabular}
\end{table}

\begin{figure}[!ht]
\centering
\includegraphics[width=0.8\linewidth]{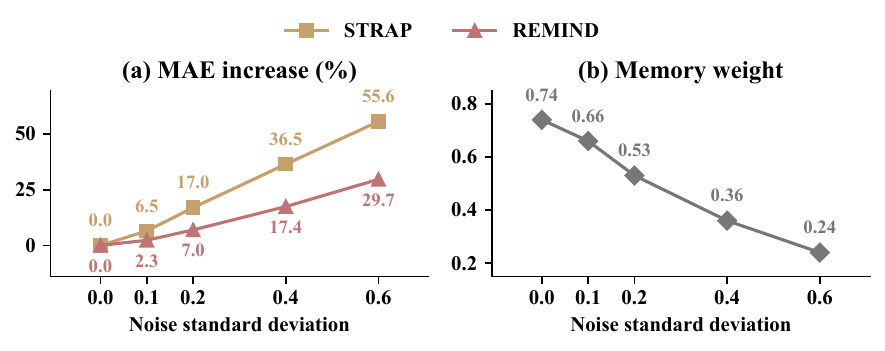}
\caption{Robustness to memory corruption on ENERGY.}
\label{fig:memory-corruption}
\end{figure}

To examine whether REMIND adapts when retrieved memory becomes less informative, we add Gaussian noise of increasing strength to retrieved samples at test time on ENERGY. Noise strength is measured by its standard deviation in normalized data units. Table~\ref{tab:memory-corruption} reports forecasting errors for REMIND and STRAP, while Fig.~\ref{fig:memory-corruption} shows the relative MAE increase and REMIND's mean memory weight.

As noise increases, forecasting errors rise for both methods, but REMIND degrades more gradually. Its memory weight also decreases, indicating that the gate reduces direct reuse of corrupted historical targets. These observations are consistent with the intended adaptation: REMIND places less weight on retrieved targets as their quality declines, helping limit the effect of memory corruption on forecasting. This further highlights the importance of assessing retrieval reliability and adaptively adjusting memory utilization.

\section{Limitations and Broader Impacts}
Although REMIND improves probabilistic forecasting under distribution shift, it still relies on the availability and quality of historical retrieval memory. Under extremely sparse, rapidly evolving, or entirely unseen scenarios, suitable historical patterns may be difficult to retrieve, which can limit the effectiveness of retrieval-guided adaptation.

For broader impacts, this work aims to improve the reliability and generalization of probabilistic spatiotemporal forecasting under non-stationary environments, which may benefit applications such as traffic management, energy systems, and hydrology. More reliable uncertainty estimation can support safer and more informed decision-making under evolving conditions. More broadly, this work highlights the potential of combining retrieval mechanisms with probabilistic modeling for robust decision support in dynamic spatiotemporal systems. However, under unseen scenarios, inaccurate or miscalibrated uncertainty estimates may still affect downstream decisions. Therefore, careful evaluation and responsible deployment remain important in real-world applications.


\clearpage
\section*{NeurIPS Paper Checklist}

\begin{enumerate}

\item {\bf Claims}
    \item[] Question: Do the main claims made in the abstract and introduction accurately reflect the paper's contributions and scope?
    \item[] Answer: \answerYes{} 
    \item[] Justification: The abstract and introduction accurately reflect the scope and contributions of the paper. The proposed retrieval-guided probabilistic framework, shift-adaptive fusion mechanism, and hierarchical Gaussian mixture are formally described in Secs.~4.2-4.5 and further analyzed in the appendix. The claims regarding forecasting accuracy, probabilistic uncertainty modeling, and generalization under distribution shift are supported by the experimental results and ablation studies in Sec.~5. 

\item {\bf Limitations}
    \item[] Question: Does the paper discuss the limitations of the work performed by the authors?
    \item[] Answer: \answerYes{} 
    \item[] Justification: The paper discusses the limitations of the proposed framework in App.~D. In particular, we note that the effectiveness of retrieval-guided forecasting may depend on the availability of relevant historical patterns under complex distribution shifts.

\item {\bf Theory assumptions and proofs}
    \item[] Question: For each theoretical result, does the paper provide the full set of assumptions and a complete (and correct) proof?
    \item[] Answer: \answerYes{} 
    \item[] Justification: The paper provides theoretical analysis of the hierarchical Gaussian mixture in App.~A, including the derivation of the unified mixture formulation and predictive variance decomposition. All formulas are explicitly defined in the main paper and appendix. The derivations are presented with the required assumptions and intermediate steps to support the probabilistic interpretation of the proposed framework.

    \item {\bf Experimental result reproducibility}
    \item[] Question: Does the paper fully disclose all the information needed to reproduce the main experimental results of the paper to the extent that it affects the main claims and/or conclusions of the paper (regardless of whether the code and data are provided or not)?
    \item[] Answer: \answerYes{} 
    \item[] Justification: The paper provides detailed descriptions of the proposed architecture, retrieval strategy, training objective, and experimental settings in Secs.~4-5 and the App.~C. Dataset splits, evaluation metrics, optimization settings, hyperparameter selection, and implementation details are explicitly reported. 

\item {\bf Open access to data and code}
    \item[] Question: Does the paper provide open access to the data and code, with sufficient instructions to faithfully reproduce the main experimental results, as described in supplemental material?
    \item[] Answer: \answerYes{} 
    \item[] Justification: The datasets used in this work are publicly available benchmark datasets. The paper provides an anonymous code repository in the abstract for reproducibility during the review process, and the code will be fully released upon acceptance. Detailed implementation settings, training configurations, and experimental protocols are also provided in Sec.~5 and App.~C to support faithful reproduction of the main results.

\item {\bf Experimental setting/details}
    \item[] Question: Does the paper specify all the training and test details (e.g., data splits, hyperparameters, how they were chosen, type of optimizer) necessary to understand the results?
    \item[] Answer: \answerYes{} 
    \item[] Justification: The paper provides detailed descriptions of the proposed architecture, retrieval strategy, training objective, and experimental settings in Secs.~4-5 and the App.~C. Dataset splits, evaluation metrics, optimization settings, hyperparameter selection, and implementation details are explicitly reported. 

\item {\bf Experiment statistical significance}
    \item[] Question: Does the paper report error bars suitably and correctly defined or other appropriate information about the statistical significance of the experiments?
    \item[] Answer: \answerYes{} 
    \item[] Justification: As stated in Sec.~5.1, all experimental results are averaged over three runs with different random seeds. Standard deviations are reported in tables where space permits, reflecting variability caused by random initialization and training stochasticity. In addition, box plots with distribution quantiles are provided in Sec.~5 to further illustrate the statistical behavior of the proposed method.

\item {\bf Experiments compute resources}
    \item[] Question: For each experiment, does the paper provide sufficient information on the computer resources (type of compute workers, memory, time of execution) needed to reproduce the experiments?
    \item[] Answer: \answerYes{} 
    \item[] Justification: The paper reports the experimental environment in Sec.~5.1, including the use of PyTorch 2.4.0 and an NVIDIA GeForce RTX A4000 GPU. App.~B further reports the computational efficiency of different retrieval strategies and discusses their scalability under different retrieval granularities.
    
\item {\bf Code of ethics}
    \item[] Question: Does the research conducted in the paper conform, in every respect, with the NeurIPS Code of Ethics \url{https://neurips.cc/public/EthicsGuidelines}?
    \item[] Answer: \answerYes{} 
    \item[] Justification: The research presented in this paper conforms to the NeurIPS Code of Ethics. The work uses publicly available datasets in Sec.~5, preserves author anonymity during submission, and does not involve human subjects or sensitive personal data. The paper also discusses limitations and broader impacts of the proposed framework in App.~D.

\item {\bf Broader impacts}
    \item[] Question: Does the paper discuss both potential positive societal impacts and negative societal impacts of the work performed?
    \item[] Answer: \answerYes{} 
    \item[] Justification: The paper discusses broader impacts in Sec.~1, which highlights the potential benefits of improving uncertainty modeling for spatiotemporal forecasting applications such as traffic management, energy systems, and hydrology. A more comprehensive discussion of both positive and potential negative impacts is further provided in App.~D, including the potential influence of uncertainty estimation quality under unseen scenarios and the importance of careful evaluation in real-world applications.
    
\item {\bf Safeguards}
    \item[] Question: Does the paper describe safeguards that have been put in place for responsible release of data or models that have a high risk for misuse (e.g., pre-trained language models, image generators, or scraped datasets)?
    \item[] Answer: \answerNA{} 
    \item[] Justification: The proposed method is a probabilistic spatiotemporal forecasting framework and does not involve high-risk generative models, scraped datasets, or sensitive personal data that would require special safeguards for release.

\item {\bf Licenses for existing assets}
    \item[] Question: Are the creators or original owners of assets (e.g., code, data, models), used in the paper, properly credited and are the license and terms of use explicitly mentioned and properly respected?
    \item[] Answer: \answerYes{} 
    \item[] Justification: The paper uses publicly available benchmark datasets and properly cites the original sources of the datasets and baseline methods. The usage of these assets follows their publicly available terms and research usage policies. Relevant references and dataset descriptions are provided in Sec.~5 and the App.C.

\item {\bf New assets}
    \item[] Question: Are new assets introduced in the paper well documented and is the documentation provided alongside the assets?
    \item[] Answer: \answerYes{} 
    \item[] Justification: An anonymized code repository with usage documentation is provided in the abstract during the review process, and the full codebase and related documentation will be publicly released upon acceptance.

\item {\bf Crowdsourcing and research with human subjects}
    \item[] Question: For crowdsourcing experiments and research with human subjects, does the paper include the full text of instructions given to participants and screenshots, if applicable, as well as details about compensation (if any)? 
    \item[] Answer: \answerNA{} 
    \item[] Justification: This paper does not involve crowdsourcing experiments or research with human subjects.

\item {\bf Institutional review board (IRB) approvals or equivalent for research with human subjects}
    \item[] Question: Does the paper describe potential risks incurred by study participants, whether such risks were disclosed to the subjects, and whether Institutional Review Board (IRB) approvals (or an equivalent approval/review based on the requirements of your country or institution) were obtained?
    \item[] Answer: \answerNA{} 
    \item[] Justification: This paper does not involve research with human subjects or crowdsourcing experiments, and therefore no IRB approval is required.

\item {\bf Declaration of LLM usage}
    \item[] Question: Does the paper describe the usage of LLMs if it is an important, original, or non-standard component of the core methods in this research? Note that if the LLM is used only for writing, editing, or formatting purposes and does \emph{not} impact the core methodology, scientific rigor, or originality of the research, declaration is not required.
    \item[] Answer: \answerNA{} 
    \item[] Justification: The core methodology and experiments in this paper do not involve LLMs as important, original, or non-standard components.

\end{enumerate}

\end{document}